\documentclass[11pt]{article}

\usepackage[preprint]{acl}

\usepackage{times}
\usepackage{latexsym}

\usepackage[T1]{fontenc}

\usepackage[utf8]{inputenc}

\usepackage{microtype}

\usepackage{inconsolata}

\usepackage{graphicx}
 \usepackage{microtype}
\usepackage{algorithm}
\usepackage{algorithmic}
\usepackage[most]{tcolorbox}

\usepackage{amsmath,amsfonts,bm}

\def\eqref#1{equation~\ref{#1}}

\def\1{\bm{1}}

\DeclareMathAlphabet{\mathsfit}{\encodingdefault}{\sfdefault}{m}{sl}
\SetMathAlphabet{\mathsfit}{bold}{\encodingdefault}{\sfdefault}{bx}{n}

\usepackage[table]{xcolor} 
\usepackage{subcaption}
\usepackage{colortbl}      
\usepackage{hyperref}
\usepackage{url}
\usepackage{multirow}
\usepackage{graphicx}
\usepackage{hyperref}
\usepackage{xcolor}
\usepackage{longtable}

\newcommand{\bea}{\begin{eqnarray}} 
\newcommand{\eea}{\end{eqnarray}}

\newcommand{\eg}{\textit{e.g., }}
\newcommand{\ie}{\textit{i.e., }}

\usepackage{booktabs}
\usepackage{multirow}

\usepackage{fancyhdr}

\usepackage{bm}
\usepackage{textcomp}
\usepackage{footnote}
\usepackage{enumerate}

\usepackage{color}
\usepackage{xcolor}
\usepackage{colortbl}

\usepackage{nicematrix}
\usepackage{booktabs}
\usepackage{makecell}
\usepackage{diagbox}
\usepackage{wrapfig}

\usepackage{float}

\usepackage{marvosym}

\usepackage[most]{tcolorbox}  
\tcbset{
  mdquote/.style={
    enhanced,
    breakable,
    colback=gray!5,
    colframe=gray!50,
    left=4pt,
    right=4pt,
    top=4pt,
    bottom=4pt,
    sharp corners,
    before skip=6pt,
    after skip=6pt,
    borderline west={2pt}{0pt}{gray!70}, 
    boxrule=0pt,
  }
}

\usepackage{graphicx}
\graphicspath{{figures/}}

\makeatletter
\def\input@path{{tables/}{appendices/}}
\makeatother

\newcommand{\rowstrut}{\rule[-0.6ex]{0pt}{2.4ex}}

\usepackage{dsfont}

\usepackage{threeparttable}

\usepackage{enumitem}

\usepackage{tablefootnote}

\newcommand\newfootnote[1]{%
  \begingroup
  \renewcommand\thefootnote{}\footnote{#1}%
  \addtocounter{footnote}{-1}%
  \endgroup
}
\title{FAITH: \\Factuality Alignment through Integrating Trustworthiness and Honestness}

\author{
\textbf{Xiaoning Dong\textsuperscript{$\ast$1,2}}~~~
  \textbf{Chengyan Wu\textsuperscript{$\ast$3}}~~~
\textbf{Yajie Wen\textsuperscript{3}}~~~
\textbf{Yu Chen\textsuperscript{1}}~~~
\\
\textbf{Yun Xue\textsuperscript{2}}~~~
\textbf{Jing Zhang\textsuperscript{$\ddagger$4}}~~~
\textbf{Wei Xu\textsuperscript{$\ddagger$1,2}}~~~
\textbf{Bolei Ma\textsuperscript{5}}
\vspace{2.5pt}
\\
\small{\textsuperscript{1}Tsinghua University, Institute for Interdisciplinary Information Sciences}~
\small{\textsuperscript{2}Shanghai Qi Zhi Institute}~\\
\small{\textsuperscript{3}South China Normal University, School of Electronic Science and Engineering}\\
\small{\textsuperscript{4}Guangzhou Richstone Data Technologies Co., Ltd.}~
\small{\textsuperscript{5}LMU Munich \& Munich Center for Machine Learning}
\vspace{2pt}
\\
\small{
\texttt{\{chengyan.wu, yajiewen, xueyun\}@m.scnu.edu.cn, \{dongxn20, chenyu23\}@mails.tsinghua.edu.cn }
}\\
\small{
\texttt{
zhangjing@richstonedt.com,  weixu@mail.tsinghua.edu.cn, bolei.ma@lmu.de}
}
}

\begin{document}
\maketitle
\begin{abstract}
Large Language Models (LLMs) can generate factually inaccurate content even if they have corresponding knowledge, which critically undermines their reliability. Existing approaches attempt to mitigate this by incorporating uncertainty in QA prompt during training, but these numerical scores lack the semantic richness for LLM to properly understand its internal states of trustworthiness and honestness, leading to insufficient factuality alignment. We introduce \textbf{FAITH} (\textbf{F}actuality \textbf{A}lignment through \textbf{I}ntegrating \textbf{T}rustworthiness and \textbf{H}onestness), a post-training framework for factuality alignment that integrates natural-language uncertainty signals with external knowledge. Specifically, we augment training datasets by computing confidence scores and semantic entropy from LLM outputs and mapping them into a knowledge state quadrant that describes the model's internal knowledge possession (trustworthiness) and answering behaviors (honestness) in natural language.\newfootnote{$^\ast$Equal contributions.}\newfootnote{$^\ddagger$Corresponding authors.}
Based on this enhanced data, we design a reward function that considers both correctness and uncertainty signals, and fine-tune the LLM using the Proximal Policy Optimization (PPO) algorithm. To further mitigate weakly grounded responses, we design a retrieval-augmented module that retrieves relevant external passages, 
improving the consistency between internal and external knowledge representations. Extensive experiments on four knowledge-intensive benchmarks demonstrate that FAITH enhances the factual accuracy and truthfulness of LLMs.\footnote{Code: \url{https://github.com/xndong/FAITH}}
\end{abstract}

\section{Introduction}
\label{sec:introduction}
Large Language Models (LLMs), such as GPT-4~\citep{DBLP:journals/corr/abs-2303-08774}, Llama3~\citep{dubeyLlama3Herd2024} and DeepSeek-v3~\citep{deepseek-aiDeepSeekV3TechnicalReport2024}, have demonstrated impressive performance across a broad range of natural language processing tasks. Despite these advances, growing evidence shows that LLMs may generate outputs that are fluent but factually incorrect or fabricated, a phenomenon commonly known as hallucination~\citep{DBLP:journals/corr/abs-2104-14839,DBLP:journals/csur/JiLFYSXIBMF23}. Such hallucinations pose substantial risks in knowledge-intensive and high-stakes domains, including legal, educational, and clinical applications~\citep{alkaissi2023artificial,DBLP:conf/acl/WangKMLSKH23}.

A concerning type of hallucination emerges when the model possesses the necessary knowledge but fails to articulate it correctly. This disconnect between internal knowledge and external expression, often termed the \textit{know–tell} gap~\citep{DBLP:journals/corr/abs-2206-05802,DBLP:journals/tmlr/Li0WS0ZC0YL0YWW25}, not only undermines the model’s ability to convey truthful information but also manifests as inconsistency of factual expression, where the model may produce an incorrect response in one instance yet a correct one in another~\citep{DBLP:conf/emnlp/ManakulLG23,DBLP:conf/iclr/0002WSLCNCZ23}.  

In this work, we propose to post-train LLMs for enhancing factuality. Our study identifies several limitations in recent endeavors (\citealp[][\textit{inter alia}]{tian2024finetuning,DBLP:conf/acl/TaoYDXC0GSD24,xue-etal-2025-ualign, sun-etal-2025-divide}): (1) while these work introduce uncertainty for factual alignment, they directly use the numerical values into question-answering prompts during training, which lack semantic richness and are difficult for LLMs to understand and exploit for factuality-aligned expression; (2) they employ binary reward function in policy training, which simply focuses on whether the response is correct or not while ignoring to consider the confidence of LLM's response (\ie uncertainty), potentially encouraging guessing; and (3) they neglect the use of external knowledge, leaving potentially incorrect responses unrectified. 

To address these limitations, we introduce \textbf{FAITH} (\textbf{F}actuality \textbf{A}lignment through \textbf{I}ntegrating \textbf{T}rustworthiness and \textbf{H}onestness), a post-training framework designed for factuality alignment in LLMs. 
FAITH incorporates three key designs: 
(1) When augment in-domain training datasets, beyond estimating the uncertainties (via consistency and semantic entropy) of LLMs for each question in datasets, we map these numerical values into a knowledge state quadrant~\citep{liangLearningTrustYour2024} where each knowledge state is expressed in natural language and defined along two dimensions: \textit{knowledge possession} (trustworthiness) and \textit{answering behavior} (honestness). Unlike opaque numerical uncertainty values, incorporating knowledge states into QA prompt during training provides LLMs with semantically rich and interpretable guidance.   
(2) For policy optimization, we design a fine-grained reward function that consider both the correctness of response and LLM's uncertainty, providing more informative feedback than a binary reward, encouraging the policy model to align its outputs with their knowledge states. 
(3) To further improve reliability, we construct a vector database over the Wikipedia corpus~\citep{karpukhin-etal-2020-dense} and train a RAG model that retrieves external knowledge from the database as contextual input to rectify potentially incorrect responses generated by the policy model.

Through FAITH, we enhance LLM factuality in terms of precision and truthfulness. Extensive experiments show that FAITH consistently outperforms five recent strong baselines on three in-domain and one out-of-domain dataset. For example, on Llama3-8B, FAITH achieves 74.26\% precision and 45.73\% truthfulness on in-domain datasets, and 67.99\% precision and 34.03\% truthfulness on the out-of-domain dataset. Similar gains are observed on Mistral-7B-v0.1, demonstrating that FAITH’s effectiveness generalizes across both models and datasets.

In summary, our contributions are as follows:
\begin{enumerate}[leftmargin=*]
    \item We introduce FAITH, a novel post-training framework for factuality alignment. FAITH advances the factuality alignment by its semantically rich knowledge state quadrant, fine-grained reward function, and employing external knowledge to ground LLM's response.
    \item We conduct extensive experiments demonstrating that FAITH consistently outperforms strong baselines, with performance gains generalizing across multiple datasets and models. Meanwhile, we provide ablations to assess the contribution of each component.
    \item We provide in-depth analyses of FAITH, including the impact of different knowledge state estimation strategies on inference performance and training-time scaling behavior with varying numbers of sampled responses $K$.
\end{enumerate}

\section{Preliminary}
\label{sec:preliminary}

\paragraph{Problem Definition.} 

\begin{table}[htbp]
    \centering
    \small
    \begin{tabular}{cl}
        \toprule
        \multicolumn{1}{c}{\textbf{Abbr.}} & \textbf{Explanation} \\
        \midrule
        \textbf{KC} & Known and answered correctly  \\
        \textbf{KI} & Known but answered incorrectly \\
        \textbf{KR} & Known but refused to answer \\
        \textbf{UC} & Unknown but answered correctly \\
        \textbf{UI} & Unknown but answered incorrectly \\
        \textbf{UR} & Unknown and refused to answer \\
        \bottomrule
    \end{tabular}
    \caption{Categories of model output.}
    \label{tab:preliminary}%
\end{table}

We consider a standard \textit{open-domain} setting, where a language model \( LLM \) is given a factual question \( q \) and generates a short-form answer \( a \sim LLM(\cdot \mid q) \) with probability $LLM(a\mid q)$. The answer is expected to be concise and factually correct, but in practice may suffer from failure of factuality due to uneven knowledge possession and the gap between knowledge and expression, while the autoregressive generation paradigm inevitably produces responses by sampling from token distributions.

Following prior work~\citep{xue-etal-2025-ualign}, we categorize model outputs into six types based on knowledge possession and response correctness, as summarized in Table~\ref{tab:preliminary}. 
The research scope of this work is to encourage more instances of \textbf{KC} and \textbf{UR} (See \S~\ref{subsec:exp_setup} for evaluation metrics). 

\section{The FAITH Method}
\label{methods}

\begin{figure*}[tbp] 
\centering 
\includegraphics[width=0.999\linewidth]{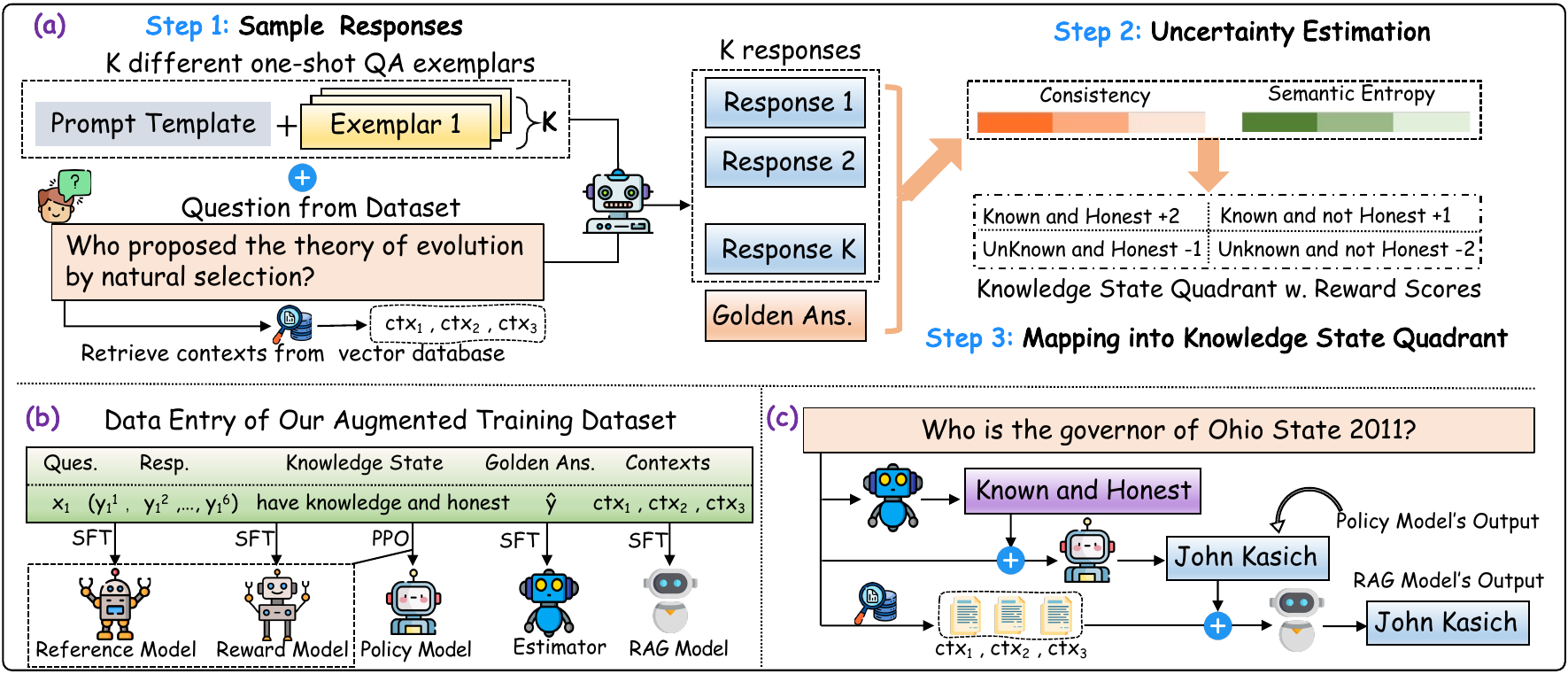} 
\caption{Illustration of the FAITH framework. Panel (a) shows the procedure for augmenting the training datasets, Panel (b) depicts model training with the augmented datasets, and Panel (c) presents the inference pipeline.} 
\label{fig:model} 
\end{figure*}
As shown in Figure~\ref{fig:model}, we introduce the framework of FAITH 
which enhances factuality alignment for LLMs. Specifically, we first augment three QA training datasets by estimating the uncertainties of LLMs for each question in these datasets and translating the numerical uncertainty values into natural-language descriptions drawn from the knowledge state quadrant (described in \S~\ref{subsec:dataset_preparation}). We then apply Proximal Policy Optimization (PPO) to finetune LLMs on the augmented datasets, guiding them to answer questions combining their knowledge state; 
meanwhile, to mitigate internal knowledge insufficiency for confidently expressing knowledge, we train a RAG model to rectify potential incorrect answers produced by policy model (described in \S~\ref{subsec:training_stage}). Finally, we conduct the inference in \S~\ref{subsec:inference_stage}.

\subsection{Training Dataset Augmentation}
\label{subsec:dataset_preparation}

As shown in Figure~\ref{fig:model}, Panel (a), we first sample responses to each question in in-domain QA datasets and then conduct uncertainty estimation, which is consistent with existing works of uncertainty estimation~\citep{xiongCanLLMsExpress2024,DBLP:conf/iclr/KuhnGF23,aichbergerImprovingUncertaintyEstimation2024,kangScalableBestofNSelection2025}. 
We do this for two purposes: (1) to measure the knowledge boundary of an LLM and locate whether a given question is within it; (2) to evaluate the honestness of an LLM when answering the given question. 

\paragraph{Sampling Responses from Training Datasets.}
We sample responses from the training split of NQ-Open~\citep{kwiatkowski-etal-2019-natural}, SciQ~\citep{welbl-etal-2017-crowdsourcing} and TriviaQA~\citep{joshi-etal-2017-triviaqa} datasets, and each dataset $\mathcal{D}$ contains a set of $N$ question-answer pairs $\{(x_i, \hat y_i)\}_{i=1}^{N}$, where $x_i$ and $\hat y_i$ represent the $i$-th question and golden answer in $\mathcal{D}$, respectively. Specifically, for each question $q_i \in \mathcal{D}$, we prompt $q_i$ with $K$ different one-shot exemplars and obtain $K$ responses, denoted as $Y_i=\{y_i^k\}_{k=1}^{K}$. We set  $K=6$ in the main experiment, same as in baseline UAlign~\citep{xue-etal-2025-ualign}. 
For fair comparison
, we adopt the same temperature \texttt{T=0.2} with UAlign and also randomly preserve half of data entries in NQ-Open and TriviaQA datasets while preserve all entries in SciQ dataset. 

\paragraph{Uncertainty Estimation.}
For the uncertainty estimation of generative LLMs, we employ consistency and semantic entropy. Consistency serves as an accuracy-based measure of confidence, and it reflects the accuracy of the generated $K$ candidate responses~\citep{xiongCanLLMsExpress2024}, computed as follows:
\begin{equation}
Consistency(x_i) =\frac{1}{K} \sum_{i=1}^K \mathds{1}\left\{y_i^k=\hat{y_i}\right\}.
\label{eq:consistency}
\end{equation}
The semantic entropy (SE), on the other hand, captures uncertainty from the semantic dispersion of generated answers, determining the likelihood of each meaning $c$ rather than each generated sequence $y_i ^ k$~\citep{DBLP:conf/iclr/KuhnGF23}. 
It addresses the limitations of prior approaches, which are often affected by response length or by semantically identical answers expressed in different surface forms. Semantic entropy is defined as:

\begin{equation}
\small
\begin{split}
SE(x_i) &= -\sum_c p(c \mid x_i) \log p(c \mid x_i) \\
&
= -\sum_c 
        \left(
            \Big( \sum_{y_i^k \in c} p(y_i^k \mid x_i) \Big)
            \log \Big[ \sum_{y_i^k \in c} p(y_i^k \mid x_i) \Big]
        \right)
\end{split}
\label{eq:semantic_entropy}
\end{equation}
By now, we augment each data entry in $\mathcal{D}$ from $(x_i, \hat y_i)$ to $(x_i, \hat y_i, Y_i, Consistency(x_i), SE(x_i))$. 
However, we argue that the numerical values of uncertainty, which will be used in QA training prompt (\eg “\texttt{Who starred in an officer and a gentleman \#\#\# Conf: 0.833 \#\#\# Entro: -0.}” in~\citet{xue-etal-2025-ualign}), cannot effectively guide LLM to understand and recognize its knowledge boundary, as raw numbers lack semantic meaning. 

\paragraph{Knowledge State Mapping.} To this end, we map consistency scores and semantic entropy onto our defined knowledge state quadrant (described below), thus rendering otherwise opaque numerical uncertainty values in semantically rich natural language descriptions. 
Specifically, we describe knowledge states of LLM based on two factors: 
\textit{knowledge possession} and \textit{answer behavior}, and this results in a quadrant with four knowledge states expressed in natural language: 
(i) \textbf{Have knowledge and honesty (KH)}, 
(ii) \textbf{Have knowledge but not honesty (K$\lnot$H)}, 
(iii) \textbf{Not have knowledge but honesty ($\lnot$KH)}, and 
(iv) \textbf{Not have knowledge and not honesty ($\lnot$K$\lnot$H)}.

We quantify the {knowledge possession} of LLMs for a given question $x_i$ through consistency defined in Equation~\ref{eq:consistency}, and the indicator function $\mathds{1}$ is \textit{Positive-Recall Exact Match (PREM)}, where $\text{PREM}(y_i ^ k, \hat{y_i}) = 1$ if $y_i \in \hat y_i ^ k \lor \hat y_i ^ k \in y_i$, otherwise $\text{PREM}(y_i ^ k, \hat{y_i}) = 0$, which is widely-used in short-form QA. 
On the other hand, we model the {answer behavior} of LLMs via semantic entropy.

Overall, the procedure for mapping consistency and semantic entropy into the knowledge state quadrant $\mathcal{S}$ is defined as follows (We explain the details of knowledge states in Appendix \ref{cognitive_state_mapping}.):

\begin{equation}
\small
\begin{split}
s_i &= \text{KnowledgeState}(x_i) \\&=
\begin{cases}
\text{KH}, & \text{if }  \text{Consistency}(x_i) > 0 \ \text{and}\ SE(x_i) = 0, \\[3pt]
\text{K}\lnot\text{H}, & \text{if } \text{Consistency}(x_i) > 0 \ \text{and}\ SE(x_i) \neq 0, \\[3pt]
\lnot\text{KH}, & \text{if } \text{Consistency}(x_i) = 0 \ \text{and}\ SE(x_i) = 0, \\[3pt]
\lnot\text{K}\lnot\text{H}, & \text{otherwise.}
\end{cases}
\end{split}
\label{state_cog}
\end{equation}

The knowledge state formulation (Eq. \ref{state_cog}) allows us to  characterize LLMs in terms of trustworthiness (knowledge posession) and honestness (answer behavior), and we finally augment each data entry 
from $(x_i, \hat y_i)$ to $(x_i, \hat y_i, Y_i, s_i)$.

\subsection{Training Stage of FAITH}
\label{subsec:training_stage}
We aim to leverage both internal and external knowledge to enhance the expression of existing knowledge, bridging the gap between knowing and telling. 
To this end, as shown in Figure~\ref{fig:model}, Panel (b), we first train a policy model using PPO to align LLM's responses with its internal knowledge states. We then train a RAG model to correct potentially incorrect responses by incorporating external knowledge. 
Finally, we introduce a knowledge state estimator that eliminates the need for sampling multiple responses during inference, thereby improving efficiency. 
All prompt templates used in training stage are provided in Appendix~\ref{prompt-template}. 

\paragraph{Reference Model Training.}
We start from a pretrained base model and obtain a reference model $\pi_{\mu}$ through supervised fine-tuning (SFT). Specifically, the model is fine-tuned using pairs of the form $(\texttt{prompt}(x_i, s_i); \hat{y}_i)$, where the prompt incorporates both the question $x_i$ and the knowledge state $s_i$, and $\hat{y}_i$ is the golden answer. By fine-tuning the base model with these curated input–output pairs, the reference model establishes a foundation for subsequent policy optimization.

\paragraph{Reward Model Training.}
\label{para:reward_model_training}
To align generation with knowledge state, we train a reward model with parameter $\theta$ to evaluate the generated response combined with the knowledge state. 
Different from existing binary reward $r_i \in \{0,1\}$ which only focuses on whether the response is correct or not and ignores how confident the correct response is~\citep{yaoAreReasoningModels2025,kirichenkoAbstentionBenchReasoningLLMs,xue-etal-2025-ualign}, we propose a fine-grained reward function to focus on both the correctness of response and the uncertainty. Specifically, we propose a combined reward function:
\begin{equation}
\small
\begin{split}
R_{\mathrm{FAITH}}(x_i, y_i^k, \hat y_i, s_i)
    &= R_{\text{correctness}}(y_i^k, \hat y_i)
       + R_{\text{uncertainty}}(s_i) \\
    &= \mathds{1}_{y_i^k \equiv \hat y_i}
       + R_{\text{uncertainty}}(s_i).
\end{split}
\label{eq:reward_func}
\end{equation}
where $s_i=\text{KnowledgeState}(x_i) \in \mathcal{S}$ and the $R_{\text{uncertainty}} \in \{+2, +1, -1, -2\}$ is defined by the following rules in terms of its knowledge state $s_i$ in $\mathcal{S}$:
\[
\begin{aligned}
+2 &\rightarrow \text{KH}, \qquad
+1 &\rightarrow \text{K}\lnot\text{H}, \\
-1 &\rightarrow \lnot\text{KH}, \qquad
-2 &\rightarrow \lnot\text{K}\lnot\text{H}.
\end{aligned}
\]
We parameterize the reward function into a reward model $RM_\theta$. Specifically, given a dataset $\mathcal{D}$ containing multiple tuples $(x_i, y_i^k, s_i, r_i^k)$, where $r_i^k$ is the reward value, the reward model minimizes the multi-class cross-entropy:
\begin{equation}
\mathcal{L}_{\theta} = - \mathbb{E}_{(x_i, y_i^k,  s_i, r_i^k) \sim \mathcal{D}} 
\left[ \log p_{\theta}(r_i^k \mid x_i, y_i^k, s_i) \right].
\label{eq:rm_loss}
\end{equation}
Our fine-grained reward model provides more informative feedback than a binary reward, encouraging the policy model to align its generated responses with their knowledge state, where uncertainty is expressed in natural language form rather than numeric scores.
\paragraph{Policy Model Training}
Similar to reinforcement learning from human feedback (RLHF)~\citep{DBLP:conf/nips/Ouyang0JAWMZASR22}, we employ PPO with a KL-divergence penalty to optimize LLMs for factuality alignment. 
Specifically, given a question $x_i$ paired with its knowledge state $s_i$, both the reference model $\pi_\mu$ and the policy model $\pi_\phi$ generate responses, while the reward model $RM_\theta$ evaluates the factual reliability of a generated response $\tilde y_i$ in terms of correctness and uncertainty (i.e., knowledge state). 
The training objective is to optimize $\pi_\phi$ to maximize the expected reward:
\begin{equation}
\small
\begin{aligned}
\arg\max_{\pi_{\phi}} \ 
&\mathbb{E}_{x \sim \mathcal{D},\ s \sim \text{KnowledgeState}(x),\ 
      \tilde y \sim \pi_{\phi}(x, s)} \Big[
    \underbrace{RM_\theta(x, \tilde y, s)}_{\text{reward}} \\
&\qquad\qquad\qquad\qquad
    - \beta\,
    \underbrace{\operatorname{KL}\!\left[\pi_{\mu}(x)\ \|\ \pi_{\phi}(x, s)\right]}_{\text{penalty}}
\Big].
\end{aligned}
\end{equation}

\paragraph{RAG Model Training.}
\label{subsec:rag}
We train a RAG model $\pi_{rag}$ to leverage external knowledge to rectify potentially incorrect answers produced by the policy model. 
To this end, we first build a vector database over the Wikipedia corpus~\citep{karpukhin-etal-2020-dense} using the BAAI General Embedding model\footnote{BGE-base-en-v1.5 :  \href{https://huggingface.co/BAAI/bge-base-en-v1.5}{https://huggingface.co/BAAI/bge-base-en-v1.5}}. We employ IndexIVFPQ in Facebook AI Similarity Search (FAISS)~\citep{DBLP:journals/tbd/JohnsonDJ21} as the retriever to perform similarity search. 
For each question $x_i \in \mathcal{D}$, the retriever returns the top-3 most semantically relevant passages, denoted as $ctx_i = \{context_i ^ {j}\}_{j=1} ^ {3}$, which are used as  context in prompt. Accordingly, we augment the training dataset entries from $(x_i, \hat y_i, Y_i, s_i)$ to $(x_i, \hat y_i, Y_i, s_i, ctx_i)$. 
Finally, we perform retrieval-augmented
fine-tuning (RAFT)~\citep{zhangRAFTAdaptingLanguage2024} of an LLM as the rectifier, using training pairs of the form $(\text{prompt}(x_i, s_i, \bar y_i, ctx_i); \hat y_i)$, where $\bar y_i$ is randomly selected from $K$ responses in $Y_i$.

\paragraph{Knowledge State Estimator Training.} 
To improve inference efficiency, we additionally train a knowledge state estimator that directly predicts the LLM’s knowledge state $s_i$ for a given question $x_i \in \mathcal{D}$. Since we represent knowledge possession and answer behavior within a knowledge state quadrant, the estimator is formulated as a four-class classification task. 

Specifically, given the augmented training dataset $\mathcal{D} = {(x_i, \hat y_i, Y_i, s_i)}_{i=1}^N$ described in \S~\ref{subsec:dataset_preparation}, we perform supervised fine-tuning of an LLM to serve as the knowledge state estimator, using pairs of the form $(\texttt{prompt}(x_i); s_i)$, where the prompt incorporates the question $x_i$ and the target label is its knowledge state $s_i$. The estimator is parameterized by $\tau$, and its SFT objective is defined as:
\begin{equation}
\mathcal{L}_{\tau} = - \mathbb{E}_{(x_i, s_i) \sim \mathcal{D}} \left[ \log p_{\tau}(s_i \mid x_i) \right].
\label{eq:sft_loss}
\end{equation}
This design enables the estimator to obtain a knowledge state in a single forward pass, rather than relying on sampling $K$ responses and computing consistency and semantic entropy. 
We provide empirical evaluations of the estimator’s impact on model performance in \S~\ref{para:estimator_sampleK}.

\subsection{Inference Stage of FAITH}
\label{subsec:inference_stage}
We employ the policy model $\pi_{\phi}$, the estimator model $Est_{\tau}$, and the RAG model $\pi_{rag}$ to perform factuality-enhanced question answering. Specifically, as shown in Figure~\ref{fig:model}, Panel (c), given a question $x$, we first predict its knowledge state $s$ in the knowledge state quadrant using the estimator model: $s = \text{Est}_\tau(x)$. We then prompt the policy model $\pi\phi$ with $(x, s)$ to generate the answer $\tilde y = \pi_\phi(\text{prompt}(x, s))$. Finally, we apply the RAG model to further rectify the answer produced by the policy model: $\tilde {y} ^ \star = \pi_{rag}(\text{prompt}(x, s, \tilde y, ctx_i))$, obtaining the final answer $\tilde {y} ^ \star$. 
We analyze the impact of the RAG model as a rectifier in \S~\ref{para:impact_of_rag}.  
All prompt templates used during inference are identical to those employed in the training stage.

\section{Experiments}
\label{sec:exp}
\subsection{Experimental Setup}
\label{subsec:exp_setup}
\paragraph{Datasets.}
For training, we adopt the same widely used QA datasets as prior works in FAITH to ensure fair comparison: 
(1) TriviaQA~\citep{joshi-etal-2017-triviaqa}, where questions are from various topics and authored by trivia enthusiasts with evidence documents.
(2) SciQ~\citep{welbl-etal-2017-crowdsourcing}, which focuses on question answering in the scientific domain; and (3) NQ-Open~\citep{kwiatkowski-etal-2019-natural}, consisting of Google search queries paired with annotated short-form answers. For evaluation, we use the test splits of these three datasets as in-domain benchmarks, and employ WebQuestions~\citep{DBLP:conf/emnlp/BerantCFL13} as an out-of-domain dataset to assess the generalization capability of our approach. Detailed descriptions and statistics of all datasets are provided in Appendix~\ref{dataset detail}.

\paragraph{Evaluation Metrics.}
Consistent with baselines, we employ Precision (\textit{Prec.}) and Truthfulness (\textit{Truth.}) as evaluation metrics. Precision measures the proportion of correctly answered questions among all known questions, reflecting an LLM’s ability to accurately articulate its known knowledge. Truthfulness is defined as the proportion of correctly answered known questions plus correctly refused unknown questions over all questions. Further details are provided in Appendix~\ref{evaluation metric}.
\paragraph{Baselines.}
We compare FAITH against the following baselines spanning three categories: 
prompt-based, SFT-based, and RL-based methods.

\textbf{(1) ICL-CoT}~\citep{DBLP:conf/nips/Wei0SBIXCLZ22} is a prompt-based approach 
that employs few-shot exemplars with chain-of-thought reasoning steps to elicit more 
accurate answers from LLMs without any parameter updates.

\textbf{(2) SFT} fine-tunes LLMs by minimizing the negative log-likelihood of 
ground-truth answers conditioned on the input questions, serving as a standard 
supervised learning baseline.

\textbf{(3) RL-DPO} follows~\citet{DBLP:conf/nips/LinGOXLY024} to construct a 
factuality preference dataset and applies direct preference optimization to improve 
the factual accuracy of LLMs.

\textbf{(4) Divide-then-Align (DTA)}~\citep{sun-etal-2025-divide} is a multi-objective 
training framework for retrieval-augmented LLMs that combines DPO loss, SFT loss, and 
a boundary classification loss to align model behavior with explicit knowledge boundary 
constraints.

\textbf{(5) UAlign}~\citep{xue-etal-2025-ualign} leverages uncertainty estimation to 
train LLMs to accurately express their factual knowledge boundaries, particularly for 
questions they cannot consistently answer correctly.

\textbf{(6) Context-DPO}~\citep{bi-etal-2025-context} and \textbf{(7) 
InFact}~\citep{cohen-etal-2025-infact} are two additional state-of-the-art baselines 
included for broader comparison.

To assess the benefit of retrieval augmentation, we further implement \textbf{SFT$_{rag}$} 
and \textbf{UAlign$_{rag}$} by integrating the same RAG module used in FAITH into the 
respective baselines.

\paragraph{Training Setup.}
We implement our approach on Llama3-8B~\citep{dubeyLlama3Herd2024} and Mistral-7B-v0.1~\citep{jiang2023mistral7b}, applying LoRA for parameter-efficient fine-tuning. Full training details are provided in Appendix~\ref{training_details}.

\begin{table*}[t]
  \centering
    \resizebox{\linewidth}{!}{
    \begin{tabular}{ccccccccccc}
    \toprule
    \multicolumn{1}{c}{\multirow{2}[4]{*}{\textbf{Method}}} & \multicolumn{2}{c}{\textbf{TVQA (ID)}} & \multicolumn{2}{c}{\textbf{SciQ (ID)}} & \multicolumn{2}{c}{\textbf{NQ-Open (ID)}} & \multicolumn{2}{c}{\textbf{Average (ID)}}& \multicolumn{2}{c}{\textbf{WebQ-QA (OOD)}} \\
\cmidrule{2-11}          & \multicolumn{1}{c}{\textit{Prec.} $\uparrow$} & \multicolumn{1}{c}{\textit{Truth.} $\uparrow$} & \multicolumn{1}{c}{\textit{Prec.} $\uparrow$} & \multicolumn{1}{c}{\textit{Truth.} $\uparrow$} & \multicolumn{1}{c}{\textit{Prec.} $\uparrow$} & \multicolumn{1}{c}{\textit{Truth.} $\uparrow$} & \multicolumn{1}{c}{\textit{Prec.} $\uparrow$} & \multicolumn{1}{c}{\textit{Truth.} $\uparrow$}& \multicolumn{1}{c}{\textit{Prec.} $\uparrow$} & \multicolumn{1}{c}{\textit{Truth.} $\uparrow$} \\ 
\specialrule{0.1em}{0pt}{0pt} 
    \rowcolor[rgb]{0.91, 0.91, 0.91} \multicolumn{11}{c}{\rowstrut Llama3-8B} \\
\specialrule{0pt}{0ex}{.05ex}  
    \multicolumn{1}{l}{\textbf{ICL-CoT}} &   66.68&  53.37 &  72.34 &  45.90 &  57.34 &  23.60          &  65.45 &40.95          &   65.97    &  30.85                    \\
    \multicolumn{1}{l}{\textbf{SFT}} &   70.80 &   52.57 &    72.18   & 45.40        &41.41  & 16.57  &61.46   &38.18        &    66.46   &   31.18               \\

    \multicolumn{1}{l}{\textbf{SFT$_{rag}$}} & \multicolumn{1}{c}{71.49} & \multicolumn{1}{c}{53.04} & \multicolumn{1}{c}{73.23} & \multicolumn{1}{c}{45.97} & \multicolumn{1}{c}{42.64} & \multicolumn{1}{c}{17.58} & \multicolumn{1}{c}{62.45} & \multicolumn{1}{c}{38.86}&\multicolumn{1}{c}{66.93} & \multicolumn{1}{c}{31.85} \\

    \multicolumn{1}{l}{\textbf{RL-DPO}} &  72.08   & 53.96   &   71.23     &  44.20  &  49.65   & 19.18     & 64.32   & 39.11      &  65.99     &   32.41            \\
        \multicolumn{1}{l}{\textbf{DTA\footnotemark}} &    43.99   & 31.73      &   --    &   --    &   56.72    &  21.12   &  --     & --  & 61.24 &30.71\\
    \multicolumn{1}{l}{\textbf{UAlign}} & \multicolumn{1}{c}{79.14} & \multicolumn{1}{c}{57.04} & \multicolumn{1}{c}{76.44} & \multicolumn{1}{c}{48.00} & \multicolumn{1}{c}{56.60} & \multicolumn{1}{c}{26.09} & \multicolumn{1}{c}{70.72} & \multicolumn{1}{c}{43.71}&\multicolumn{1}{c}{66.88} & \multicolumn{1}{c}{33.01} \\
        \multicolumn{1}{l}{\textbf{UAlign$_{sft}$}} & \multicolumn{1}{c}{78.76} & \multicolumn{1}{c}{56.68} & \multicolumn{1}{c}{75.87} & \multicolumn{1}{c}{47.65} & \multicolumn{1}{c}{56.02} & \multicolumn{1}{c}{25.49} & \multicolumn{1}{c}{70.22} & \multicolumn{1}{c}{43.27}&\multicolumn{1}{c}{66.12} & \multicolumn{1}{c}{32.58} \\
        \multicolumn{1}{l}{\textbf{UAlign$_{rag}$}} & \multicolumn{1}{c}{79.35} & \multicolumn{1}{c}{57.24} & \multicolumn{1}{c}{77.47} & \multicolumn{1}{c}{48.25} & \multicolumn{1}{c}{56.78} & \multicolumn{1}{c}{26.10} & \multicolumn{1}{c}{ 71.20} & \multicolumn{1}{c}{43.86}&\multicolumn{1}{c}{67.01} & \multicolumn{1}{c}{33.21} \\
        \multicolumn{1}{l}{\textbf{Context-DPO}} & \multicolumn{1}{c}{79.76} & \multicolumn{1}{c}{57.26} & \multicolumn{1}{c}{77.32} & \multicolumn{1}{c}{48.62} & \multicolumn{1}{c}{56.82} & \multicolumn{1}{c}{26.02} & \multicolumn{1}{c}{71.30} & \multicolumn{1}{c}{43.97 }&\multicolumn{1}{c}{66.75} & \multicolumn{1}{c}{33.12} \\
        \multicolumn{1}{l}{\textbf{InFact}} & \multicolumn{1}{c}{80.11} & \multicolumn{1}{c}{57.76} & \multicolumn{1}{c}{76.94} & \multicolumn{1}{c}{48.21} & \multicolumn{1}{c}{56.93} & \multicolumn{1}{c}{26.11} & \multicolumn{1}{c}{71.33} & \multicolumn{1}{c}{44.03}&\multicolumn{1}{c}{66.92} & \multicolumn{1}{c}{33.44} \\

    \midrule
    \multicolumn{1}{l}{\textbf{FAITH (ours)}} & \multicolumn{1}{c}{\textbf{84.19}} & \multicolumn{1}{c}{\textbf{60.69}} & \multicolumn{1}{c}{\textbf{80.61}} & \multicolumn{1}{c}{\textbf{49.99}} & \multicolumn{1}{c}{\textbf{58.13}} & \multicolumn{1}{c}{\textbf{27.58}} & \multicolumn{1}{c}{\textbf{74.26}} & \multicolumn{1}{c}{\textbf{45.73}}& \multicolumn{1}{c}{\textbf{67.99}} & \multicolumn{1}{c}{\textbf{34.03}} \\
    \multicolumn{1}{l}{\textbf{FAITH$_{sft+ppo}$}} & \multicolumn{1}{c}{82.95} & \multicolumn{1}{c}{59.80} & \multicolumn{1}{c}{80.29} & \multicolumn{1}{c}{49.70} & \multicolumn{1}{c}{57.99} & \multicolumn{1}{c}{26.52} & \multicolumn{1}{c}{73.79} & \multicolumn{1}{c}{45.69}& \multicolumn{1}{c}{67.31}& \multicolumn{1}{c}{33.75} \\
       \multicolumn{1}{l}{\textbf{FAITH$_{sft}$}} & \multicolumn{1}{c}{81.24} & \multicolumn{1}{c}{58.77} & \multicolumn{1}{c}{78.85} & \multicolumn{1}{c}{48.65} & \multicolumn{1}{c}{56.72} & \multicolumn{1}{c}{26.21} & \multicolumn{1}{c}{72.27} & \multicolumn{1}{c}{44.54}& \multicolumn{1}{c}{66.94} & \multicolumn{1}{c}{33.10} \\
\specialrule{0.1em}{0pt}{0pt} 
    \rowcolor[rgb]{0.91, 0.91, 0.91} \multicolumn{11}{c}{\rowstrut Mistral-7B-v0.1} \\
\specialrule{0pt}{0ex}{.05ex}  
    \multicolumn{1}{l}{\textbf{ICL-CoT}} &   76.73  &  54.78   &  71.87     & 44.20   &  54.47     &  18.22   & 67.69      &  39.06  &   53.43   &   35.76              \\
    \multicolumn{1}{l}{\textbf{SFT}} &   74.57  &   54.77    &  65.85    &   42.50      &   50.82   &14.42   &63.74      &37.08     &   52.24   &   34.33            \\

    \multicolumn{1}{l}{\textbf{SFT$_{rag}$}} & \multicolumn{1}{c}{75.87} & \multicolumn{1}{c}{55.09} & \multicolumn{1}{c}{66.08} & \multicolumn{1}{c}{43.22} & \multicolumn{1}{c}{51.03} & \multicolumn{1}{c}{14.53} & \multicolumn{1}{c}{64.33} & \multicolumn{1}{c}{37.61}& \multicolumn{1}{c}{53.14} & \multicolumn{1}{c}{34.82} \\

    \multicolumn{1}{l}{\textbf{RL-DPO}} &   72.20   &  52.98    & 66.44   &  41.80      &50.95  &  16.42        &63.19        &37.06     &   52.01   &   33.87          \\
    \multicolumn{1}{l}{\textbf{DTA\footnotemark[\value{footnote}]}} &    41.33   &   28.78    &     --  &    --   &   41.01    &   20.49   &   --    &-- & 56.53 &23.44 \\
    \multicolumn{1}{l}{\textbf{UAlign}} & \multicolumn{1}{c}{82.10} & \multicolumn{1}{c}{59.05} & \multicolumn{1}{c}{73.21} & \multicolumn{1}{c}{46.70} & \multicolumn{1}{c}{54.17} & \multicolumn{1}{c}{19.64} & \multicolumn{1}{c}{70.82} & \multicolumn{1}{c}{41.79}& \multicolumn{1}{c}{56.47} & \multicolumn{1}{c}{37.02} \\
    \multicolumn{1}{l}{\textbf{UAlign$_{sft}$}} & \multicolumn{1}{c}{81.07} & \multicolumn{1}{c}{56.47} & \multicolumn{1}{c}{72.45} & \multicolumn{1}{c}{45.87} & \multicolumn{1}{c}{43.32} & \multicolumn{1}{c}{21.55} & \multicolumn{1}{c}{65.61} & \multicolumn{1}{c}{41.30}&\multicolumn{1}{c}{55.34} & \multicolumn{1}{c}{36.87} \\

    \multicolumn{1}{l}{\textbf{UAlign$_{rag}$}} & \multicolumn{1}{c}{82.74} & \multicolumn{1}{c}{59.55} & \multicolumn{1}{c}{74.22} & \multicolumn{1}{c}{46.89} & \multicolumn{1}{c}{54.47} & \multicolumn{1}{c}{20.01} & \multicolumn{1}{c}{70.48} & \multicolumn{1}{c}{42.15}&\multicolumn{1}{c}{56.97} & \multicolumn{1}{c}{37.79} \\
    \multicolumn{1}{l}{\textbf{Context-DPO}} & \multicolumn{1}{c}{84.32} & \multicolumn{1}{c}{59.51} & \multicolumn{1}{c}{75.13} & \multicolumn{1}{c}{47.89} & \multicolumn{1}{c}{47.21} & \multicolumn{1}{c}{22.51} & \multicolumn{1}{c}{ 68.89} & \multicolumn{1}{c}{43.30}&\multicolumn{1}{c}{57.32} & \multicolumn{1}{c}{37.99 } \\
    \multicolumn{1}{l}{\textbf{InFact}} & \multicolumn{1}{c}{83.44} & \multicolumn{1}{c}{59.32} & \multicolumn{1}{c}{75.32} & \multicolumn{1}{c}{48.03} & \multicolumn{1}{c}{47.01} & \multicolumn{1}{c}{22.33} & \multicolumn{1}{c}{ 68.59} & \multicolumn{1}{c}{43.23}&\multicolumn{1}{c}{56.98} & \multicolumn{1}{c}{37.73} \\

    \midrule
    \multicolumn{1}{l}{\textbf{FAITH (ours)}} &  \textbf{87.20}     &   \textbf{60.72}    &   81.42    &   51.40    &   \textbf{48.05}    & \textbf{23.91}    &   72.22    & \textbf{45.34} &   \textbf{58.04}    &   40.43   \\
    \multicolumn{1}{l}{\textbf{FAITH$_{sft+ppo}$}} & \multicolumn{1}{c}{87.00} & \multicolumn{1}{c}{60.58} & \multicolumn{1}{c}{\textbf{83.68}} & \multicolumn{1}{c}{\textbf{51.80}} & \multicolumn{1}{c}{46.60} & \multicolumn{1}{c}{23.19} & \multicolumn{1}{c}{\textbf{72.43}} & \multicolumn{1}{c}{45.19} & 55.48  & 38.65\\
    \multicolumn{1}{l}{\textbf{FAITH$_{sft}$}} &  86.51  &  60.24  & 82.88 & 51.30 &  46.16 &  22.96 &   71.85    & 44.83  &  51.99 & \textbf{41.77}\\
    \bottomrule

    \end{tabular}%
    }
      \caption{\textbf{Precision and Truthfulness of FAITH (ours)} vs. strong baselines on in-domain (ID) and out-of-domain (OOD) QA datasets. 
  The Average (ID) column denotes the average performance on all three ID datasets.
  The subscript ``sft'' denotes ablation results with only supervised fine-tuning (SFT), excluding the PPO and RAG (if applicable) module.  Similarly, ``sft+ppo'' denotes results with SFT and PPO, but excluding the RAG module. 
  All results are reported in percentages.} 
  \label{tab:main-results}
\end{table*}%

\footnotetext{For DTA with Llama3, we directly evaluate the released checkpoint, whereas for DTA with Mistral, we fine-tune Mistral-7B-Instruct on the released training data for Llama3 in a transfer setting. Meanwhile, since DTA requires augmented QA datasets with RAG context and SciQ's augmented version was not released, its results on SciQ are unavailable (denoted as “–” in the table).}

\subsection{Main Results}

We evaluate the effectiveness of our factuality alignment framework FAITH against strong baselines with experimental results presented in Table \ref{tab:main-results}. From the table, the following key findings are:
\paragraph{(1) FAITH achieves state-of-the-art performance, outperforming advanced baselines.}
As shown in Table~\ref{tab:main-results}, FAITH consistently surpasses five baselines on three in-domain and one out-of-domain dataset.
For instance, on Llama3-8B model, FAITH achieves an overall precision of 74.26\% and truthfulness 45.73\% on in-domain datasets, and it attains precision of 67.99\% and truthfulness of 34.03\% on WebQuestions dataset. 
We observe similar performance superiority on Mistral-7B model, with the exception of precision on NQ-Open, demonstrating that the effectiveness of FAITH generalizes across models and datasets.

\paragraph{(2) Natural-language knowledge states are more effective than numerical uncertainty values in guiding knowledge-boundary-aware question answering.}
To assess the effectiveness of our knowledge-state-quadrant design, we compare it against numerical uncertainty values. Specifically, we construct a variant of UAlign by eliminating its policy optimization stage and retaining the remaining SFT stage, i.e., we apply SFT with prompts containing numerical uncertainty values. We keep all other settings unchanged. Similarly, we implement FAITH with SFT only, where the model is prompted with natural-language knowledge state drawn from the knowledge-state quadrant. Their performance is reported in Table~\ref{tab:main-results} under \textbf{UAlign$_{sft}$} and \textbf{FAITH$_{sft}$}, respectively. 

Evaluation shows that replacing numerical uncertainty values with semantically rich knowledge states in natural language yields clear gains in guiding LLMs to understand their knowledge boundary and answer questions accordingly. For instance, on Llama3-8B, FAITH with SFT outperforms UAlign with SFT by 2.05\% in precision and 1.27\% in truthfulness on average, with even larger improvements observed on Mistral-7B. 
We attribute these improvements to LLMs’ preference for semantically meaningful labels (e.g., “known”, “honest”) that better convey knowledge boundary. In contrast to fitting abstract numerical values, LLMs more readily interpret and leverage natural language as guidance, enabling knowledge-boundary-aware question answering.
\paragraph{(3) Reinforcement learning with our proposed reward function improves performance.} We examine the impact of reward function design by comparing the correctness-based binary reward used in UAlign with our fine-grained reward function in Eq.~\ref{eq:reward_func}. For example, on Llama3-8B, applying PPO with binary reward yields average gains of 0.7\% in precision and 0.44\% in truthfulness over SFT\footnote{Calculated as the difference between the metric values reported under \textbf{UAlign} and \textbf{UAlign$_{sft}$} in Table~\ref{tab:main-results}} on three in-domain datasets, whereas FAITH, applying PPO with our reward function, achieves larger improvements of 1.52\% in precision and 1.15\% in truthfulness, which demonstrates the effectiveness of the fine-grained reward function in incentivizing LLM's generation from both correctness and uncertainty. 

\paragraph{(4) Retrieval-Augmented Fine-Tuning aligns policy model outputs with external knowledge by rectifying potential errors.}
As shown in Table~\ref{tab:main-results}, comparing the values under \textbf{FAITH} with \textbf{FAITH$_{sft+ppo}$}, we observe consistent performance improvements across both LLMs, except for SciQ on Mistral-7B. This demonstrates that incorporating external knowledge enhances the truthfulness of LLM's responses. Besides, we manually inspect the corrections made by RAG model to the policy model outputs. Interestingly, some rectifications fail, even altering correct answers into incorrect ones, though such cases are rare. We provide an in-depth analysis of such cases in \S~\ref{para:impact_of_rag}.

\subsection{Analysis and Discussion}
\label{subsec:analysis}
\paragraph{Performance of the RAG Model on Post-Hoc Correction to Policy Model Outputs.}
\label{para:impact_of_rag}

\begin{figure}[htbp]
    \includegraphics[width=\linewidth]{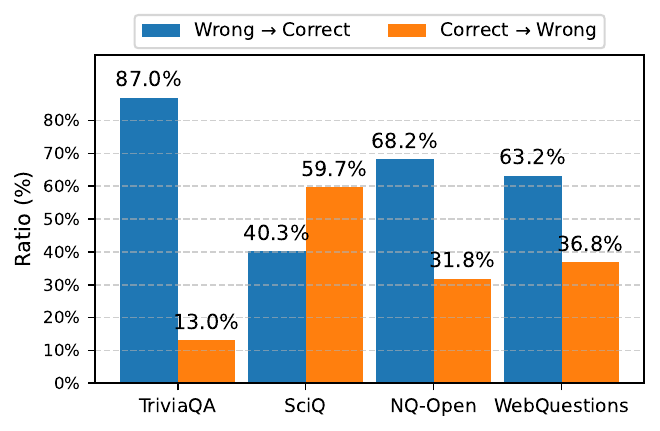} 
    \centering
    \caption{Ratios of on four datasets.}
    \label{fig:rag_error} 
\end{figure}

For this analysis, we conduct a statistical study on both in-domain (TriviaQA, SciQ, NQ-Open) and out-of-domain (WebQuestions) datasets, with results summarized in Figure~\ref{fig:rag_error}. Specifically, we examine the responses produced by the policy model that are subsequently modified by the RAG model, and compute the proportion of cases where an incorrect response is corrected into a correct one versus the reverse. We find that the proportion of correct rectifications consistently exceeds that of erroneous rectifications across all datasets except SciQ. Notably, on TriviaQA, 87\% of the policy model outputs modified by the RAG model are corrected successfully, demonstrating that effectively leveraging external knowledge can compensate for the limitations of relying solely on internal knowledge.

\paragraph{Calculate Knowledge State: Estimator vs. Sample $K$ Responses.}
\label{para:estimator_sampleK}
We further investigate the impact of different knowledge state estimation strategies on model performance during inference, comparing model-based estimation and sampling-based estimation. 
The model-based approach corresponds to our trained estimator model, whereas the sampling-based approach follows a two-stage pipeline: first sampling $K$ responses using prompts with different one-shot examples, and then computing the knowledge state with Eq.~\ref{state_cog}. For this analysis, we set $K=6$, consistent with the training stage. 
We present the evaluation results for precision and truthfulness in Table~\ref{tab:self-eval-results}, and we observe that sampling-based estimation yields slightly higher precision and truthfulness in most cases. 

These findings indicate the distribution of the knowledge states can be captured by a trained LLM, while also highlighting a trade-off between efficiency and performance: sampling provides better performance with interpretable uncertainty measures, whereas model-based estimation avoids $K$ rounds of inference with only minimal performance degradation.

\begin{table*}[htbp]
  \centering
  \resizebox{\linewidth}{!}{
    \begin{tabular}{ccccccccccc}
    \toprule
    \multicolumn{1}{c}{\multirow{2}[4]{*}{\textbf{Method}}} & \multicolumn{2}{c}{\textbf{TVQA (ID)}} & \multicolumn{2}{c}{\textbf{SciQ (ID)}} & \multicolumn{2}{c}{\textbf{NQ-Open (ID)}} & \multicolumn{2}{c}{\textbf{Average (ID)}}& \multicolumn{2}{c}{\textbf{WebQ-QA (OOD)}} \\
\cmidrule{2-11}          & \multicolumn{1}{c}{\textit{Prec.} $\uparrow$} & \multicolumn{1}{c}{\textit{Truth.} $\uparrow$} & \multicolumn{1}{c}{\textit{Prec.} $\uparrow$} & \multicolumn{1}{c}{\textit{Truth.} $\uparrow$} & \multicolumn{1}{c}{\textit{Prec.} $\uparrow$} & \multicolumn{1}{c}{\textit{Truth.} $\uparrow$} & \multicolumn{1}{c}{\textit{Prec.} $\uparrow$} & \multicolumn{1}{c}{\textit{Truth.} $\uparrow$}& \multicolumn{1}{c}{\textit{Prec.} $\uparrow$} & \multicolumn{1}{c}{\textit{Truth.} $\uparrow$} \\ 
\specialrule{0.1em}{0pt}{0pt} 
    \rowcolor[rgb]{0.91, 0.91, 0.91} \multicolumn{11}{c}{\rowstrut Llama3-8B} \\

\specialrule{0pt}{0ex}{.05ex}  
    \multicolumn{1}{l}{\textbf{Estimator}} &  82.95 &59.80 &80.29 &49.70 &57.99 &26.52 &73.79 &45.69 &67.31 &33.75                  \\
    \multicolumn{1}{l}{\textbf{Sample-based}} &  \textbf{83.99}  &  \textbf{59.86}  &   \textbf{83.20}  &   \textbf{51.50}    & \textbf{58.23} & \textbf{26.93} & \textbf{75.14}  &  \textbf{46.10}    & \textbf{67.85}      &     \textbf{34.07}             \\
        \rowcolor[rgb]{0.91, 0.91, 0.91} \multicolumn{11}{c}{\rowstrut Mistral-7B-v0.1} \\

\specialrule{0pt}{0ex}{.05ex}  
    \multicolumn{1}{l}{\textbf{Estimator}} &   
    87.00 &60.58 &83.68 &51.80 &\textbf{46.60} &\textbf{23.19} &\textbf{72.43} &45.19 &55.48 &38.65                      \\
    \multicolumn{1}{l}{\textbf{Sample-based}} & \textbf{87.43} & \textbf{60.88} & \textbf{84.01} & \textbf{52.00} & 45.66 & 22.71 & 72.37  & \textbf{45.20}   &  \textbf{55.70}  &  \textbf{38.80}   \\
    \bottomrule
    \end{tabular}%
    }
\caption{Performance comparison between model-based and sampling-based knowledge state estimation ($K=6$). Results are reported on precision and truthfulness across all datasets and models.}

  \label{tab:self-eval-results}
\end{table*}%

\begin{figure}[htbp]
\centering 
\includegraphics[width=1.02\linewidth]{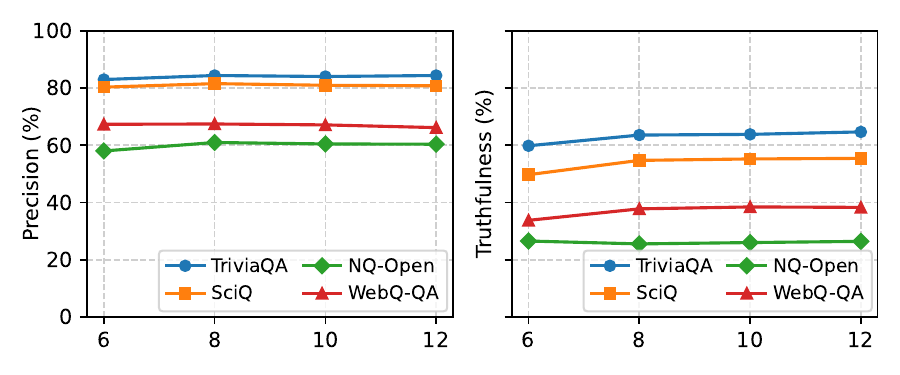} 
\caption{Training-time scaling with different numbers of sampled responses ($K$) on Llama3-8B.} 
\label{fig:training_scaling} 
\end{figure}

\paragraph{Training-time Scaling: the Impact of Number of Sampled Responses.} 
We study the training-time scaling behavior, i.e., how the number of sampled responses $K$ used during data augmentation influences the training performance of the policy model. The default $K$ we use is 6 in the main framework for efficiency. Here, specifically, we increase $K$ from \texttt{6} to \texttt{8}, \texttt{10}, and \texttt{12} during data augmentation, resulting in augmented datasets that differ only in the values of $K$. These datasets are then used to train both estimator and policy model (including reference model and reward model). Finally, we evaluate the trained models and compare the performance. 
As shown in Figure~\ref{fig:training_scaling}, increasing $K$ beyond 6 does not yield noticeable improvements in either precision or truthfulness. 
This suggests that sampling 
$K=6$ responses during data augmentation is already sufficient and effective to capture the distribution of the model’s knowledge state, while also keeping the efficiency. 
In other words, while larger $K$ values slightly expand the coverage of sampled responses, they do not translate into significant gains in downstream performance, indicating minor effects. 
The detailed numerical results are provided in Table~\ref{tab:self-eval-K-results} in the Appendix \ref{app:numerical results}. 

Finally, we present case studies in Appendix~\ref{app:case_study}.

\section{Related Work}
\label{sec:related_work}

\paragraph{Factuality Alignment.} 
To enhance LLM factuality, prior work has explored training-free strategies such as external knowledge augmentation~\citep{DBLP:conf/icml/KandpalDRWR23,DBLP:conf/emnlp/JiangXGSLDYCN23}, decoding methods~\citep{DBLP:conf/iclr/ChuangXLKGH24}, and self-consistency techniques grounded in uncertainty estimation~\citep{DBLP:journals/corr/abs-2207-05221,DBLP:conf/emnlp/TianMZSRYFM23}. More recently, post-training approaches, including SFT and policy optimization, have been applied to further improve factuality~\citep{tian2024finetuning,DBLP:conf/acl/TaoYDXC0GSD24,sun-etal-2025-divide,xuRejectionImprovesReliability2024,xue-etal-2025-ualign,chenLearningReasonFactuality2025}. Our work falls into this category, where we propose to map numerical uncertainty values into natural-language knowledge states and design a new reward function to enhance the model’s expression of its knowledge. 

\paragraph{Uncertainty Estimation.}
Uncertainty Estimation (UE) has long been studied in machine learning domain, including NLP, with vast majority of previous work focusing on discriminative tasks, such as sentiment analysis~\citep{DBLP:conf/emnlp/XiaoLBNSM22}. Specifically, the entropy of the predictive posterior and the negative predictive posterior probability of the most probable answer are used to quantify uncertainty in predictions~\citep{lakshminarayanan2017simple,bakmanMARSMeaningAwareResponse2024}. 
However, LLMs pose new challenges for uncertainty estimation due to their generative paradigm. 
Recent studies have extended UE to generative models, introducing heuristic or probabilistic metrics such as entropy-based scoring~\citep{malinin2020uncertainty}, semantic entropy that accounts for meaning equivalence in multiple generations~\citep{DBLP:conf/iclr/KuhnGF23}, and similarity-based methods applied in tasks like machine translation~\citep{DBLP:journals/tacl/FomichevaSYBGFA20,lin-etal-2022-towards}. Other works explore black-box UE by leveraging sampled outputs~\citep{DBLP:conf/acl/Chen024,DBLP:conf/emnlp/ManakulLG23}, or prompt-based approaches where models verbalize their own confidence~\citep{DBLP:journals/corr/abs-2207-05221}. Training-based methods have also been proposed to enhance linguistic self-assessments of uncertainty~\citep{DBLP:journals/tmlr/LinHE22}. 

\paragraph{Retrieval-Augmented Generation.}
Retrieval-Augmented Generation (RAG) has become a widely adopted strategy for mitigating the limitations of parametric knowledge in LLMs by grounding responses in external evidence~\citep{DBLP:journals/corr/abs-2002-08909,DBLP:conf/eacl/IzacardG21,DBLP:conf/emnlp/ZhongWMPC23}. 
Unlike knowledge editing methods that directly modify model parameters~\citep{DBLP:conf/emnlp/YaoWT0LDC023,DBLP:conf/cikm/XuZZLL00WY0C024}, RAG enables models to dynamically access updated information without retraining, reducing the risk of catastrophic forgetting. 
However, recent studies reveal that contradictions may arise when retrieved knowledge conflicts with internal representations~\citep{DBLP:conf/emnlp/MeiLWBC24,DBLP:conf/acl/NiBGC24}. To mitigate this, retrieval-based methods increasingly leverage structured repositories such as knowledge graphs for more reliable grounding~\citep{DBLP:journals/corr/abs-2306-04136,DBLP:journals/corr/abs-2305-13669}. Other approaches enhance factuality by incorporating retrieval into inference-time interventions, including memory-augmented architectures~\citep{DBLP:conf/nips/LiGK22} and entity-level embedding integration~\citep{DBLP:conf/naacl/KangBH22,DBLP:conf/iclr/JongZFSC22}.
Generally, RAG provides a scalable way to continuously integrate knowledge, offering advantages over task-specific parameter editing.

\section{Conclusion}
We present a post-training framework, called FAITH, for factuality alignment in LLMs. Our approach estimates uncertainty and translates the numerical values into natural-language knowledge states that measure the knowledge possession and answering behavior of LLMs. Meanwhile, we design a fine-grained reward function to incentivize both correctness and uncertainty of LLM's response. Finally, we introduce a trained RAG model to rectify potentially incorrect responses generated by policy model. 
Experiments show that FAITH substantially outperforms recent baselines in truthfulness and precision. We hope this work contributes to building more faithful and factual LLMs as part of the broader community effort.

\section*{Limitations}
\label{app:limitation}
\paragraph{Reward Function Design.} 
Our reward function is derived from heuristic rules that are straightforward to formulate and intuitively easy to interpret. In practice, we observe that this design works well empirically and provides meaningful guidance for aligning model behavior. However, the current formulation lacks rigorous theoretical guarantees, leaving room for future work to establish a stronger theoretical foundation for its effectiveness.

\paragraph{Computational Overhead.} During dataset construction, we sample $K$ responses and build a vector database. At inference time, our pipeline first uses $Est_\tau$ to estimate the knowledge state $s$, then applies the policy model $\pi_\phi$ to generate an answer, and finally employs $\pi_{rag}$ for rectification. Even without rectifica4tion, two model inferences are required, rather than a single end-to-end pass. 
For future work, we plan to explore more efficient approaches for cognitive-state estimation, such as lightweight estimators derived from LLM internal representations~\citep{zhuLLMAlreadyKnows2025}.

\paragraph{Unexplored Aspects of RAG Effectiveness.} Our current study does not investigate how the quality of the data used to build the vector database affects FAITH’s performance. For example, on the SciQ dataset with Mistral-7B, incorporating external knowledge does not improve correction effectiveness, which may be related to the quality of the retrieved context. In addition, we have not explored more effective ways of leveraging external knowledge for rectification, such as integrating RAFT directly the SFT stage and accordingly applying PPO training on top of the RAFT-enhanced model, rather than additionally train a RAG model.

\section*{Ethical Considerations}

This research was conducted in accordance with the ACM Code of Ethics. 
The datasets \cite{joshi-etal-2017-triviaqa, welbl-etal-2017-crowdsourcing, kwiatkowski-etal-2019-natural, DBLP:conf/emnlp/BerantCFL13} that we use are publicly available. 
We report only aggregated results in the main paper.
We did not share any Personally Identifiable Data with this paper. 
The project itself does not raise any risks.

\section*{Acknowledgments}
CW and YX are supported in part by the Guangdong Basic and Applied Basic Research Foundation under Grant 2023A1515011370, the National Natural Science Foundation of China (32371114), the Characteristic Innovation Projects of Guangdong Colleges and Universities (No. 2018KTSCX049), and the Guangdong Provincial Key Laboratory (No. 2023B1212060076). XD and WX are supported in part by the National Key
R\&D Program of China 2023YFC3304802 and National Natural Science Foundation of China (NSFC)
Grant U2268202 and 62176135.

\bibliography{custom}

@inproceedings{liangLearningTrustYour2024,
    title = "Learning to Trust Your Feelings: Leveraging Self-awareness in {LLM}s for Hallucination Mitigation",
    author = "Liang, Yuxin  and
      Song, Zhuoyang  and
      Wang, Hao  and
      Zhang, Jiaxing",
    editor = "Yu, Wenhao  and
      Shi, Weijia  and
      Yasunaga, Michihiro  and
      Jiang, Meng  and
      Zhu, Chenguang  and
      Hajishirzi, Hannaneh  and
      Zettlemoyer, Luke  and
      Zhang, Zhihan",
    booktitle = "Proceedings of the 3rd Workshop on Knowledge Augmented Methods for NLP",
    month = aug,
    year = "2024",
    address = "Bangkok, Thailand",
    publisher = "Association for Computational Linguistics",
    url = "https://aclanthology.org/2024.knowledgenlp-1.4/",
    doi = "10.18653/v1/2024.knowledgenlp-1.4",
    pages = "44--58"
}

@inproceedings{
zhangRAFTAdaptingLanguage2024,
title={{RAFT}: Adapting Language Model to Domain Specific {RAG}},
author={Tianjun Zhang and Shishir G Patil and Naman Jain and Sheng Shen and Matei Zaharia and Ion Stoica and Joseph E. Gonzalez},
booktitle={First Conference on Language Modeling},
year={2024},
url={https://openreview.net/forum?id=rzQGHXNReU}
}

@inproceedings{zhuLLMAlreadyKnows2025,
    title = "The {LLM} Already Knows: Estimating {LLM}-Perceived Question Difficulty via Hidden Representations",
    author = "Zhu, Yubo  and
      Liu, Dongrui  and
      Lin, Zecheng  and
      Tong, Wei  and
      Zhong, Sheng  and
      Shao, Jing",
    editor = "Christodoulopoulos, Christos  and
      Chakraborty, Tanmoy  and
      Rose, Carolyn  and
      Peng, Violet",
    booktitle = "Proceedings of the 2025 Conference on Empirical Methods in Natural Language Processing",
    month = nov,
    year = "2025",
    address = "Suzhou, China",
    publisher = "Association for Computational Linguistics",
    url = "https://aclanthology.org/2025.emnlp-main.61/",
    doi = "10.18653/v1/2025.emnlp-main.61",
    pages = "1160--1176",
    ISBN = "979-8-89176-332-6"
}

@inproceedings{
xuRejectionImprovesReliability2024,
title={Rejection Improves Reliability: Training {LLM}s to Refuse Unknown Questions Using {RL} from Knowledge Feedback},
author={Hongshen Xu and Zichen Zhu and Situo Zhang and Da Ma and Shuai Fan and Lu Chen and Kai Yu},
booktitle={First Conference on Language Modeling},
year={2024},
url={https://openreview.net/forum?id=lJMioZBoR8}
}

@misc{chenLearningReasonFactuality2025,
      title={Learning to Reason for Factuality}, 
      author={Xilun Chen and Ilia Kulikov and Vincent-Pierre Berges and Barlas Oğuz and Rulin Shao and Gargi Ghosh and Jason Weston and Wen-tau Yih},
      year={2025},
      eprint={2508.05618},
      archivePrefix={arXiv},
      primaryClass={cs.CL},
      url={https://arxiv.org/abs/2508.05618}, 
}

@inproceedings{bakmanMARSMeaningAwareResponse2024,
    title = "{MARS}: Meaning-Aware Response Scoring for Uncertainty Estimation in Generative {LLM}s",
    author = "Bakman, Yavuz Faruk  and
      Yaldiz, Duygu Nur  and
      Buyukates, Baturalp  and
      Tao, Chenyang  and
      Dimitriadis, Dimitrios  and
      Avestimehr, Salman",
    editor = "Ku, Lun-Wei  and
      Martins, Andre  and
      Srikumar, Vivek",
    booktitle = "Proceedings of the 62nd Annual Meeting of the Association for Computational Linguistics (Volume 1: Long Papers)",
    month = aug,
    year = "2024",
    address = "Bangkok, Thailand",
    publisher = "Association for Computational Linguistics",
    url = "https://aclanthology.org/2024.acl-long.419/",
    doi = "10.18653/v1/2024.acl-long.419",
    pages = "7752--7767"
}

@inproceedings{
kirichenkoAbstentionBenchReasoningLLMs,
title={AbstentionBench: Reasoning {LLM}s Fail on Unanswerable Questions},
author={Polina Kirichenko and Mark Ibrahim and Kamalika Chaudhuri and Samuel Bell},
booktitle={The Thirty-ninth Annual Conference on Neural Information Processing Systems Datasets and Benchmarks Track},
year={2025},
url={https://openreview.net/forum?id=OkHC30LLpO}
}

@misc{yaoAreReasoningModels2025,
      title={Are Reasoning Models More Prone to Hallucination?}, 
      author={Zijun Yao and Yantao Liu and Yanxu Chen and Jianhui Chen and Junfeng Fang and Lei Hou and Juanzi Li and Tat-Seng Chua},
      year={2025},
      eprint={2505.23646},
      archivePrefix={arXiv},
      primaryClass={cs.CL},
      url={https://arxiv.org/abs/2505.23646}, 
}

@inproceedings{
xiongCanLLMsExpress2024,
title={Can {LLM}s Express Their Uncertainty? An Empirical Evaluation of Confidence Elicitation in {LLM}s},
author={Miao Xiong and Zhiyuan Hu and Xinyang Lu and YIFEI LI and Jie Fu and Junxian He and Bryan Hooi},
booktitle={The Twelfth International Conference on Learning Representations},
year={2024},
url={https://openreview.net/forum?id=gjeQKFxFpZ}
}

@inproceedings{aichbergerImprovingUncertaintyEstimation2024,
title={Improving Uncertainty Estimation through Semantically Diverse Language Generation},
author={Lukas Aichberger and Kajetan Schweighofer and Mykyta Ielanskyi and Sepp Hochreiter},
booktitle={The Thirteenth International Conference on Learning Representations},
year={2025},
url={https://openreview.net/forum?id=HSi4VetQLj}
}

@inproceedings{
kangScalableBestofNSelection2025,
title={Scalable Best-of-N Selection for Large Language Models via Self-Certainty},
author={Zhewei Kang and Xuandong Zhao and Dawn Song},
booktitle={The Thirty-ninth Annual Conference on Neural Information Processing Systems},
year={2025},
url={https://openreview.net/forum?id=29FRqmVQK8}
}

@inproceedings{xue-etal-2025-ualign,
    title = "{UA}lign: Leveraging Uncertainty Estimations for Factuality Alignment on Large Language Models",
    author = "Xue, Boyang  and
      Mi, Fei  and
      Zhu, Qi  and
      Wang, Hongru  and
      Wang, Rui  and
      Wang, Sheng  and
      Yu, Erxin  and
      Hu, Xuming  and
      Wong, Kam-Fai",
    editor = "Che, Wanxiang  and
      Nabende, Joyce  and
      Shutova, Ekaterina  and
      Pilehvar, Mohammad Taher",
    booktitle = "Proceedings of the 63rd Annual Meeting of the Association for Computational Linguistics (Volume 1: Long Papers)",
    month = jul,
    year = "2025",
    address = "Vienna, Austria",
    publisher = "Association for Computational Linguistics",
    url = "https://aclanthology.org/2025.acl-long.299/",
    pages = "6002--6024",
    ISBN = "979-8-89176-251-0",
}

@inproceedings{joshi-etal-2017-triviaqa,
    title = "{T}rivia{QA}: A Large Scale Distantly Supervised Challenge Dataset for Reading Comprehension",
    author = "Joshi, Mandar  and
      Choi, Eunsol  and
      Weld, Daniel  and
      Zettlemoyer, Luke",
    editor = "Barzilay, Regina  and
      Kan, Min-Yen",
    booktitle = "Proceedings of the 55th Annual Meeting of the Association for Computational Linguistics (Volume 1: Long Papers)",
    month = jul,
    year = "2017",
    address = "Vancouver, Canada",
    publisher = "Association for Computational Linguistics",
    url = "https://aclanthology.org/P17-1147/",
    doi = "10.18653/v1/P17-1147",
    pages = "1601--1611"
}

@inproceedings{welbl-etal-2017-crowdsourcing,
    title = "Crowdsourcing Multiple Choice Science Questions",
    author = "Welbl, Johannes  and
      Liu, Nelson F.  and
      Gardner, Matt",
    editor = "Derczynski, Leon  and
      Xu, Wei  and
      Ritter, Alan  and
      Baldwin, Tim",
    booktitle = "Proceedings of the 3rd Workshop on Noisy User-generated Text",
    month = sep,
    year = "2017",
    address = "Copenhagen, Denmark",
    publisher = "Association for Computational Linguistics",
    url = "https://aclanthology.org/W17-4413/",
    doi = "10.18653/v1/W17-4413",
    pages = "94--106"
}

@article{kwiatkowski-etal-2019-natural,
    title = "Natural Questions: A Benchmark for Question Answering Research",
    author = "Kwiatkowski, Tom  and
      Palomaki, Jennimaria  and
      Redfield, Olivia  and
      Collins, Michael  and
      Parikh, Ankur  and
      Alberti, Chris  and
      Epstein, Danielle  and
      Polosukhin, Illia  and
      Devlin, Jacob  and
      Lee, Kenton  and
      Toutanova, Kristina  and
      Jones, Llion  and
      Kelcey, Matthew  and
      Chang, Ming-Wei  and
      Dai, Andrew M.  and
      Uszkoreit, Jakob  and
      Le, Quoc  and
      Petrov, Slav",
    editor = "Lee, Lillian  and
      Johnson, Mark  and
      Roark, Brian  and
      Nenkova, Ani",
    journal = "Transactions of the Association for Computational Linguistics",
    volume = "7",
    year = "2019",
    address = "Cambridge, MA",
    publisher = "MIT Press",
    url = "https://aclanthology.org/Q19-1026/",
    doi = "10.1162/tacl_a_00276",
    pages = "452--466",
}

@misc{jiang2023mistral7b,
      title={Mistral 7B}, 
      author={Albert Q. Jiang and Alexandre Sablayrolles and Arthur Mensch and Chris Bamford and Devendra Singh Chaplot and Diego de las Casas and Florian Bressand and Gianna Lengyel and Guillaume Lample and Lucile Saulnier and Lélio Renard Lavaud and Marie-Anne Lachaux and Pierre Stock and Teven Le Scao and Thibaut Lavril and Thomas Wang and Timothée Lacroix and William El Sayed},
      year={2023},
      eprint={2310.06825},
      archivePrefix={arXiv},
      primaryClass={cs.CL},
      url={https://arxiv.org/abs/2310.06825}, 
}

@inproceedings{karpukhin-etal-2020-dense,
    title = "Dense Passage Retrieval for Open-Domain Question Answering",
    author = "Karpukhin, Vladimir  and
      Oguz, Barlas  and
      Min, Sewon  and
      Lewis, Patrick  and
      Wu, Ledell  and
      Edunov, Sergey  and
      Chen, Danqi  and
      Yih, Wen-tau",
    editor = "Webber, Bonnie  and
      Cohn, Trevor  and
      He, Yulan  and
      Liu, Yang",
    booktitle = "Proceedings of the 2020 Conference on Empirical Methods in Natural Language Processing (EMNLP)",
    month = nov,
    year = "2020",
    address = "Online",
    publisher = "Association for Computational Linguistics",
    url = "https://aclanthology.org/2020.emnlp-main.550/",
    doi = "10.18653/v1/2020.emnlp-main.550",
    pages = "6769--6781"
}

@article{DBLP:journals/tbd/JohnsonDJ21,
  author       = {Jeff Johnson and
                  Matthijs Douze and
                  Herv{\'{e}} J{\'{e}}gou},
  title        = {Billion-Scale Similarity Search with GPUs},
  journal      = {{IEEE} Trans. Big Data},
  volume       = {7},
  number       = {3},
  pages        = {535--547},
  year         = {2021},
  url          = {https://doi.org/10.1109/TBDATA.2019.2921572},
  doi          = {10.1109/TBDATA.2019.2921572},
  bibsource    = {dblp computer science bibliography, https://dblp.org}
}

@article{DBLP:journals/corr/abs-2303-08774,
  author       = {OpenAI},
  title        = {{GPT-4} Technical Report},
  journal      = {CoRR},
  volume       = {abs/2303.08774},
  year         = {2023},
  url          = {https://doi.org/10.48550/arXiv.2303.08774},
  doi          = {10.48550/ARXIV.2303.08774},
  eprinttype    = {arXiv},
  eprint       = {2303.08774},
  bibsource    = {dblp computer science bibliography, https://dblp.org}
}

@misc{dubeyLlama3Herd2024,
  title = {The {{Llama}} 3 {{Herd}} of {{Models}}},
  author = {Dubey, Abhimanyu and Jauhri, Abhinav and Pandey, Abhinav and Kadian, Abhishek and {Al-Dahle}, Ahmad and Letman, Aiesha and Mathur, Akhil and Schelten, Alan and Yang, Amy and Fan, Angela and Goyal, Anirudh and Hartshorn, Anthony and Yang, Aobo and Mitra, Archi and Sravankumar, Archie and Korenev, Artem et al.},
  year = {2024},
  month = aug,
  number = {arXiv:2407.21783},
  eprint = {2407.21783},
  publisher = {arXiv},
  urldate = {2024-11-13},
  archiveprefix = {arXiv}
}

@misc{deepseek-aiDeepSeekV3TechnicalReport2024,
  title = {{{DeepSeek-V3 Technical Report}}},
  author = {DeepSeek-AI and Liu, Aixin and Feng, Bei and Xue, Bing and Wang, Bingxuan and Wu, Bochao and Lu, Chengda and Zhao, Chenggang et al.},
  date = {2024-12-27},
year ={2024},
  eprint = {2412.19437},
  eprinttype = {arXiv},
  eprintclass = {cs},
  doi = {10.48550/arXiv.2412.19437},
  url = {http://arxiv.org/abs/2412.19437},
  urldate = {2025-08-08},
  pubstate = {prepublished},
  version = {1}
}

@article{DBLP:journals/corr/abs-2104-14839,
  author       = {Yi{-}Chong Huang and
                  Xiachong Feng and
                  Xiaocheng Feng and
                  Bing Qin},
  title        = {The Factual Inconsistency Problem in Abstractive Text Summarization:
                  {A} Survey},
  journal      = {CoRR},
  volume       = {abs/2104.14839},
  year         = {2021},
  url          = {https://arxiv.org/abs/2104.14839},
  eprinttype    = {arXiv},
  eprint       = {2104.14839},
  bibsource    = {dblp computer science bibliography, https://dblp.org}
}

@article{DBLP:journals/csur/JiLFYSXIBMF23,
  author       = {Ziwei Ji and
                  Nayeon Lee and
                  Rita Frieske and
                  Tiezheng Yu and
                  Dan Su and
                  Yan Xu and
                  Etsuko Ishii and
                  Yejin Bang and
                  Andrea Madotto and
                  Pascale Fung},
  title        = {Survey of Hallucination in Natural Language Generation},
  journal      = {{ACM} Comput. Surv.},
  volume       = {55},
  number       = {12},
  pages        = {248:1--248:38},
  year         = {2023},
  url          = {https://doi.org/10.1145/3571730},
  doi          = {10.1145/3571730},
  bibsource    = {dblp computer science bibliography, https://dblp.org}
}

@article{alkaissi2023artificial,
  title={Artificial hallucinations in ChatGPT: implications in scientific writing},
  author={Alkaissi, Hussam and McFarlane, Samy I},
  journal={Cureus},
  volume={15},
  number={2},
  year={2023},
  publisher={Cureus}
}

@inproceedings{DBLP:conf/acl/WangKMLSKH23,
  author       = {Yizhong Wang and
                  Yeganeh Kordi and
                  Swaroop Mishra and
                  Alisa Liu and
                  Noah A. Smith and
                  Daniel Khashabi and
                  Hannaneh Hajishirzi},
  editor       = {Anna Rogers and
                  Jordan L. Boyd{-}Graber and
                  Naoaki Okazaki},
  title        = {Self-Instruct: Aligning Language Models with Self-Generated Instructions},
  booktitle    = {Proceedings of the 61st Annual Meeting of the Association for Computational
                  Linguistics (Volume 1: Long Papers), {ACL} 2023, Toronto, Canada,
                  July 9-14, 2023},
  pages        = {13484--13508},
  publisher    = {Association for Computational Linguistics},
  year         = {2023},
  url          = {https://doi.org/10.18653/v1/2023.acl-long.754},
  doi          = {10.18653/V1/2023.ACL-LONG.754},
  bibsource    = {dblp computer science bibliography, https://dblp.org}
}

@article{DBLP:journals/tmlr/Li0WS0ZC0YL0YWW25,
title={A Survey on the Honesty of Large Language Models},
author={Siheng Li and Cheng Yang and Taiqiang Wu and Chufan Shi and Yuji Zhang and Xinyu Zhu and Zesen Cheng and Deng Cai and Mo Yu and Lemao Liu and Jie Zhou and Yujiu Yang and Ngai Wong and Xixin Wu and Wai Lam},
journal={Transactions on Machine Learning Research},
issn={2835-8856},
year={2025},
url={https://openreview.net/forum?id=FJgtVfUxLQ},
note={Survey Certification}
}

@article{DBLP:journals/corr/abs-2206-05802,
  author       = {William Saunders and
                  Catherine Yeh and
                  Jeff Wu and
                  Steven Bills and
                  Long Ouyang and
                  Jonathan Ward and
                  Jan Leike},
  title        = {Self-critiquing models for assisting human evaluators},
  journal      = {CoRR},
  volume       = {abs/2206.05802},
  year         = {2022},
  url          = {https://doi.org/10.48550/arXiv.2206.05802},
  doi          = {10.48550/ARXIV.2206.05802},
  eprinttype    = {arXiv},
  eprint       = {2206.05802},
  bibsource    = {dblp computer science bibliography, https://dblp.org}
}

@inproceedings{DBLP:conf/emnlp/ManakulLG23,
  author       = {Potsawee Manakul and
                  Adian Liusie and
                  Mark J. F. Gales},
  editor       = {Houda Bouamor and
                  Juan Pino and
                  Kalika Bali},
  title        = {SelfCheckGPT: Zero-Resource Black-Box Hallucination Detection for
                  Generative Large Language Models},
  booktitle    = {Proceedings of the 2023 Conference on Empirical Methods in Natural
                  Language Processing, {EMNLP} 2023, Singapore, December 6-10, 2023},
  pages        = {9004--9017},
  publisher    = {Association for Computational Linguistics},
  year         = {2023},
  url          = {https://doi.org/10.18653/v1/2023.emnlp-main.557},
  doi          = {10.18653/V1/2023.EMNLP-MAIN.557},
  bibsource    = {dblp computer science bibliography, https://dblp.org}
}

@inproceedings{DBLP:conf/iclr/0002WSLCNCZ23,
title={Self-Consistency Improves Chain of Thought Reasoning in Language Models},
author={Xuezhi Wang and Jason Wei and Dale Schuurmans and Quoc V Le and Ed H. Chi and Sharan Narang and Aakanksha Chowdhery and Denny Zhou},
booktitle={The Eleventh International Conference on Learning Representations },
year={2023},
url={https://openreview.net/forum?id=1PL1NIMMrw}
}

@inproceedings{DBLP:conf/nips/Ouyang0JAWMZASR22,
  author       = {Long Ouyang and
                  Jeffrey Wu and
                  Xu Jiang and
                  Diogo Almeida and
                  Carroll L. Wainwright and
                  Pamela Mishkin and
                  Chong Zhang and
                  Sandhini Agarwal and
                  Katarina Slama and
                  Alex Ray and
                  John Schulman and
                  Jacob Hilton and
                  Fraser Kelton and
                  Luke Miller and
                  Maddie Simens and
                  Amanda Askell and
                  Peter Welinder and
                  Paul F. Christiano and
                  Jan Leike and
                  Ryan Lowe},
  editor       = {Sanmi Koyejo and
                  S. Mohamed and
                  A. Agarwal and
                  Danielle Belgrave and
                  K. Cho and
                  A. Oh},
  title        = {Training language models to follow instructions with human feedback},
  booktitle    = {Advances in Neural Information Processing Systems 35: Annual Conference
                  on Neural Information Processing Systems 2022, NeurIPS 2022, New Orleans,
                  LA, USA, November 28 - December 9, 2022},
  year         = {2022},
  url          = {http://papers.nips.cc/paper\_files/paper/2022/hash/b1efde53be364a73914f58805a001731-Abstract-Conference.html},
  bibsource    = {dblp computer science bibliography, https://dblp.org}
}

@inproceedings{DBLP:conf/acl/TaoYDXC0GSD24,
  author       = {Shuchang Tao and
                  Liuyi Yao and
                  Hanxing Ding and
                  Yuexiang Xie and
                  Qi Cao and
                  Fei Sun and
                  Jinyang Gao and
                  Huawei Shen and
                  Bolin Ding},
  editor       = {Lun{-}Wei Ku and
                  Andre Martins and
                  Vivek Srikumar},
  title        = {When to Trust LLMs: Aligning Confidence with Response Quality},
  booktitle    = {Findings of the Association for Computational Linguistics, {ACL} 2024,
                  Bangkok, Thailand and virtual meeting, August 11-16, 2024},
  pages        = {5984--5996},
  publisher    = {Association for Computational Linguistics},
  year         = {2024},
  url          = {https://doi.org/10.18653/v1/2024.findings-acl.357},
  doi          = {10.18653/V1/2024.FINDINGS-ACL.357},
  bibsource    = {dblp computer science bibliography, https://dblp.org}
}

@inproceedings{DBLP:conf/emnlp/TianMZSRYFM23,
  author       = {Katherine Tian and
                  Eric Mitchell and
                  Allan Zhou and
                  Archit Sharma and
                  Rafael Rafailov and
                  Huaxiu Yao and
                  Chelsea Finn and
                  Christopher D. Manning},
  editor       = {Houda Bouamor and
                  Juan Pino and
                  Kalika Bali},
  title        = {Just Ask for Calibration: Strategies for Eliciting Calibrated Confidence
                  Scores from Language Models Fine-Tuned with Human Feedback},
  booktitle    = {Proceedings of the 2023 Conference on Empirical Methods in Natural
                  Language Processing, {EMNLP} 2023, Singapore, December 6-10, 2023},
  pages        = {5433--5442},
  publisher    = {Association for Computational Linguistics},
  year         = {2023},
  url          = {https://doi.org/10.18653/v1/2023.emnlp-main.330},
  doi          = {10.18653/V1/2023.EMNLP-MAIN.330},
  bibsource    = {dblp computer science bibliography, https://dblp.org}
}

@inproceedings{DBLP:conf/nips/Wei0SBIXCLZ22,
 author = {Wei, Jason and Wang, Xuezhi and Schuurmans, Dale and Bosma, Maarten and ichter, brian and Xia, Fei and Chi, Ed and Le, Quoc V and Zhou, Denny},
 booktitle = {Advances in Neural Information Processing Systems},
 editor = {S. Koyejo and S. Mohamed and A. Agarwal and D. Belgrave and K. Cho and A. Oh},
 pages = {24824--24837},
 publisher = {Curran Associates, Inc.},
 title = {Chain-of-Thought Prompting Elicits Reasoning in Large Language Models},
 url = {https://proceedings.neurips.cc/paper_files/paper/2022/file/9d5609613524ecf4f15af0f7b31abca4-Paper-Conference.pdf},
 volume = {35},
 year = {2022}
}

@inproceedings{DBLP:conf/aclnut/WelblLG17,
  author       = {Johannes Welbl and
                  Nelson F. Liu and
                  Matt Gardner},
  editor       = {Leon Derczynski and
                  Wei Xu and
                  Alan Ritter and
                  Tim Baldwin},
  title        = {Crowdsourcing Multiple Choice Science Questions},
  booktitle    = {Proceedings of the 3rd Workshop on Noisy User-generated Text, NUT@EMNLP
                  2017, Copenhagen, Denmark, September 7, 2017},
  pages        = {94--106},
  publisher    = {Association for Computational Linguistics},
  year         = {2017},
  url          = {https://doi.org/10.18653/v1/w17-4413},
  doi          = {10.18653/V1/W17-4413},
  bibsource    = {dblp computer science bibliography, https://dblp.org}
}

@inproceedings{DBLP:conf/icml/KandpalDRWR23,
  author       = {Nikhil Kandpal and
                  Haikang Deng and
                  Adam Roberts and
                  Eric Wallace and
                  Colin Raffel},
  editor       = {Andreas Krause and
                  Emma Brunskill and
                  Kyunghyun Cho and
                  Barbara Engelhardt and
                  Sivan Sabato and
                  Jonathan Scarlett},
  title        = {Large Language Models Struggle to Learn Long-Tail Knowledge},
  booktitle    = {International Conference on Machine Learning, {ICML} 2023, 23-29 July
                  2023, Honolulu, Hawaii, {USA}},
  series       = {Proceedings of Machine Learning Research},
  volume       = {202},
  pages        = {15696--15707},
  publisher    = {{PMLR}},
  year         = {2023},
  url          = {https://proceedings.mlr.press/v202/kandpal23a.html},
  bibsource    = {dblp computer science bibliography, https://dblp.org}
}

@inproceedings{DBLP:conf/emnlp/JiangXGSLDYCN23,
  author       = {Zhengbao Jiang and
                  Frank F. Xu and
                  Luyu Gao and
                  Zhiqing Sun and
                  Qian Liu and
                  Jane Dwivedi{-}Yu and
                  Yiming Yang and
                  Jamie Callan and
                  Graham Neubig},
  editor       = {Houda Bouamor and
                  Juan Pino and
                  Kalika Bali},
  title        = {Active Retrieval Augmented Generation},
  booktitle    = {Proceedings of the 2023 Conference on Empirical Methods in Natural
                  Language Processing, {EMNLP} 2023, Singapore, December 6-10, 2023},
  pages        = {7969--7992},
  publisher    = {Association for Computational Linguistics},
  year         = {2023},
  url          = {https://doi.org/10.18653/v1/2023.emnlp-main.495},
  doi          = {10.18653/V1/2023.EMNLP-MAIN.495},
  bibsource    = {dblp computer science bibliography, https://dblp.org}
}

@inproceedings{DBLP:conf/iclr/ChuangXLKGH24,
title={DoLa: Decoding by Contrasting Layers Improves Factuality in Large Language Models},
author={Yung-Sung Chuang and Yujia Xie and Hongyin Luo and Yoon Kim and James R. Glass and Pengcheng He},
booktitle={The Twelfth International Conference on Learning Representations},
year={2024},
url={https://openreview.net/forum?id=Th6NyL07na}
}

@article{DBLP:journals/corr/abs-2207-05221,
  author       = {Saurav Kadavath and
                  Tom Conerly and
                  Amanda Askell and
                  Tom Henighan and
                  Dawn Drain and
                  Ethan Perez and
                  Nicholas Schiefer and
                  Zac Hatfield{-}Dodds and
                  Nova DasSarma and
                  Eli Tran{-}Johnson and
                  Scott Johnston and
                  Sheer El Showk and
                  Andy Jones and
                  Nelson Elhage and
                  Tristan Hume and
                  Anna Chen and
                  Yuntao Bai and
                  Sam Bowman and
                  Stanislav Fort and
                  Deep Ganguli and
                  Danny Hernandez and
                  Josh Jacobson and
                  Jackson Kernion and
                  Shauna Kravec and
                  Liane Lovitt and
                  Kamal Ndousse and
                  Catherine Olsson and
                  Sam Ringer and
                  Dario Amodei and
                  Tom Brown and
                  Jack Clark and
                  Nicholas Joseph and
                  Ben Mann and
                  Sam McCandlish and
                  Chris Olah and
                  Jared Kaplan},
  title        = {Language Models (Mostly) Know What They Know},
  journal      = {CoRR},
  volume       = {abs/2207.05221},
  year         = {2022},
  url          = {https://doi.org/10.48550/arXiv.2207.05221},
  doi          = {10.48550/ARXIV.2207.05221},
  eprinttype    = {arXiv},
  eprint       = {2207.05221},
  bibsource    = {dblp computer science bibliography, https://dblp.org}
}

@inproceedings{
tian2024finetuning,
title={Fine-Tuning Language Models for Factuality},
author={Katherine Tian and Eric Mitchell and Huaxiu Yao and Christopher D Manning and Chelsea Finn},
booktitle={The Twelfth International Conference on Learning Representations},
year={2024},
url={https://openreview.net/forum?id=WPZ2yPag4K}
}

@InProceedings{DBLP:journals/corr/abs-2002-08909,
  title = 	 {Retrieval Augmented Language Model Pre-Training},
  author =       {Guu, Kelvin and Lee, Kenton and Tung, Zora and Pasupat, Panupong and Chang, Mingwei},
  booktitle = 	 {Proceedings of the 37th International Conference on Machine Learning},
  pages = 	 {3929--3938},
  year = 	 {2020},
  editor = 	 {III, Hal Daumé and Singh, Aarti},
  volume = 	 {119},
  series = 	 {Proceedings of Machine Learning Research},
  month = 	 {13--18 Jul},
  publisher =    {PMLR},
  url = 	 {https://proceedings.mlr.press/v119/guu20a.html}
}

@inproceedings{DBLP:conf/eacl/IzacardG21,
  author       = {Gautier Izacard and
                  Edouard Grave},
  editor       = {Paola Merlo and
                  J{\"{o}}rg Tiedemann and
                  Reut Tsarfaty},
  title        = {Leveraging Passage Retrieval with Generative Models for Open Domain
                  Question Answering},
  booktitle    = {Proceedings of the 16th Conference of the European Chapter of the
                  Association for Computational Linguistics: Main Volume, {EACL} 2021,
                  Online, April 19 - 23, 2021},
  pages        = {874--880},
  publisher    = {Association for Computational Linguistics},
  year         = {2021},
  url          = {https://doi.org/10.18653/v1/2021.eacl-main.74},
  doi          = {10.18653/V1/2021.EACL-MAIN.74},
  bibsource    = {dblp computer science bibliography, https://dblp.org}
}

@inproceedings{DBLP:conf/emnlp/ZhongWMPC23,
  author       = {Zexuan Zhong and
                  Zhengxuan Wu and
                  Christopher D. Manning and
                  Christopher Potts and
                  Danqi Chen},
  editor       = {Houda Bouamor and
                  Juan Pino and
                  Kalika Bali},
  title        = {MQuAKE: Assessing Knowledge Editing in Language Models via Multi-Hop
                  Questions},
  booktitle    = {Proceedings of the 2023 Conference on Empirical Methods in Natural
                  Language Processing, {EMNLP} 2023, Singapore, December 6-10, 2023},
  pages        = {15686--15702},
  publisher    = {Association for Computational Linguistics},
  year         = {2023},
  url          = {https://doi.org/10.18653/v1/2023.emnlp-main.971},
  doi          = {10.18653/V1/2023.EMNLP-MAIN.971},
  bibsource    = {dblp computer science bibliography, https://dblp.org}
}

@inproceedings{DBLP:conf/emnlp/YaoWT0LDC023,
  author       = {Yunzhi Yao and
                  Peng Wang and
                  Bozhong Tian and
                  Siyuan Cheng and
                  Zhoubo Li and
                  Shumin Deng and
                  Huajun Chen and
                  Ningyu Zhang},
  editor       = {Houda Bouamor and
                  Juan Pino and
                  Kalika Bali},
  title        = {Editing Large Language Models: Problems, Methods, and Opportunities},
  booktitle    = {Proceedings of the 2023 Conference on Empirical Methods in Natural
                  Language Processing, {EMNLP} 2023, Singapore, December 6-10, 2023},
  pages        = {10222--10240},
  publisher    = {Association for Computational Linguistics},
  year         = {2023},
  url          = {https://doi.org/10.18653/v1/2023.emnlp-main.632},
  doi          = {10.18653/V1/2023.EMNLP-MAIN.632},
  bibsource    = {dblp computer science bibliography, https://dblp.org}
}

@inproceedings{DBLP:conf/cikm/XuZZLL00WY0C024,
  author       = {Derong Xu and
                  Ziheng Zhang and
                  Zhihong Zhu and
                  Zhenxi Lin and
                  Qidong Liu and
                  Xian Wu and
                  Tong Xu and
                  Wanyu Wang and
                  Yuyang Ye and
                  Xiangyu Zhao and
                  Enhong Chen and
                  Yefeng Zheng},
  editor       = {Edoardo Serra and
                  Francesca Spezzano},
  title        = {Editing Factual Knowledge and Explanatory Ability of Medical Large
                  Language Models},
  booktitle    = {Proceedings of the 33rd {ACM} International Conference on Information
                  and Knowledge Management, {CIKM} 2024, Boise, ID, USA, October 21-25,
                  2024},
  pages        = {2660--2670},
  publisher    = {{ACM}},
  year         = {2024},
  url          = {https://doi.org/10.1145/3627673.3679673},
  doi          = {10.1145/3627673.3679673},
  bibsource    = {dblp computer science bibliography, https://dblp.org}
}

@inproceedings{DBLP:conf/emnlp/MeiLWBC24,
  author       = {Lingrui Mei and
                  Shenghua Liu and
                  Yiwei Wang and
                  Baolong Bi and
                  Xueqi Cheng},
  editor       = {Yaser Al{-}Onaizan and
                  Mohit Bansal and
                  Yun{-}Nung Chen},
  title        = {{SLANG:} New Concept Comprehension of Large Language Models},
  booktitle    = {Proceedings of the 2024 Conference on Empirical Methods in Natural
                  Language Processing, {EMNLP} 2024, Miami, FL, USA, November 12-16,
                  2024},
  pages        = {12558--12575},
  publisher    = {Association for Computational Linguistics},
  year         = {2024},
  url          = {https://doi.org/10.18653/v1/2024.emnlp-main.698},
  doi          = {10.18653/V1/2024.EMNLP-MAIN.698},
  bibsource    = {dblp computer science bibliography, https://dblp.org}
}

@inproceedings{DBLP:conf/acl/NiBGC24,
  author       = {Shiyu Ni and
                  Keping Bi and
                  Jiafeng Guo and
                  Xueqi Cheng},
  editor       = {Lun{-}Wei Ku and
                  Andre Martins and
                  Vivek Srikumar},
  title        = {When Do LLMs Need Retrieval Augmentation? Mitigating LLMs' Overconfidence
                  Helps Retrieval Augmentation},
  booktitle    = {Findings of the Association for Computational Linguistics, {ACL} 2024,
                  Bangkok, Thailand and virtual meeting, August 11-16, 2024},
  pages        = {11375--11388},
  publisher    = {Association for Computational Linguistics},
  year         = {2024},
  url          = {https://doi.org/10.18653/v1/2024.findings-acl.675},
  doi          = {10.18653/V1/2024.FINDINGS-ACL.675},
  bibsource    = {dblp computer science bibliography, https://dblp.org}
}

@inproceedings{DBLP:journals/corr/abs-2306-04136,
    title = "Knowledge-Augmented Language Model Prompting for Zero-Shot Knowledge Graph Question Answering",
    author = "Baek, Jinheon  and
      Aji, Alham Fikri  and
      Saffari, Amir",
    editor = "Dalvi Mishra, Bhavana  and
      Durrett, Greg  and
      Jansen, Peter  and
      Neves Ribeiro, Danilo  and
      Wei, Jason",
    booktitle = "Proceedings of the 1st Workshop on Natural Language Reasoning and Structured Explanations (NLRSE)",
    month = jun,
    year = "2023",
    address = "Toronto, Canada",
    publisher = "Association for Computational Linguistics",
    url = "https://aclanthology.org/2023.nlrse-1.7/",
    doi = "10.18653/v1/2023.nlrse-1.7",
    pages = "78--106",
}

@inproceedings{DBLP:journals/corr/abs-2305-13669,
    title = "The Knowledge Alignment Problem: Bridging Human and External Knowledge for Large Language Models",
    author = "Zhang, Shuo  and
      Pan, Liangming  and
      Zhao, Junzhou  and
      Wang, William Yang",
    editor = "Ku, Lun-Wei  and
      Martins, Andre  and
      Srikumar, Vivek",
    booktitle = "Findings of the Association for Computational Linguistics: ACL 2024",
    month = aug,
    year = "2024",
    address = "Bangkok, Thailand",
    publisher = "Association for Computational Linguistics",
    url = "https://aclanthology.org/2024.findings-acl.121/",
    doi = "10.18653/v1/2024.findings-acl.121",
    pages = "2025--2038"
}

@inproceedings{DBLP:conf/naacl/KangBH22,
  author       = {Minki Kang and
                  Jinheon Baek and
                  Sung Ju Hwang},
  editor       = {Marine Carpuat and
                  Marie{-}Catherine de Marneffe and
                  Iv{\'{a}}n Vladimir Meza Ru{\'{\i}}z},
  title        = {{KALA:} Knowledge-Augmented Language Model Adaptation},
  booktitle    = {Proceedings of the 2022 Conference of the North American Chapter of
                  the Association for Computational Linguistics: Human Language Technologies,
                  {NAACL} 2022, Seattle, WA, United States, July 10-15, 2022},
  pages        = {5144--5167},
  publisher    = {Association for Computational Linguistics},
  year         = {2022},
  url          = {https://doi.org/10.18653/v1/2022.naacl-main.379},
  doi          = {10.18653/V1/2022.NAACL-MAIN.379},
  bibsource    = {dblp computer science bibliography, https://dblp.org}
}

@inproceedings{DBLP:conf/nips/LiGK22,
 author = {Li, Zonglin and Guo, Ruiqi and Kumar, Sanjiv},
 booktitle = {Advances in Neural Information Processing Systems},
 editor = {S. Koyejo and S. Mohamed and A. Agarwal and D. Belgrave and K. Cho and A. Oh},
 pages = {21698--21710},
 publisher = {Curran Associates, Inc.},
 title = {Decoupled Context Processing for Context Augmented Language Modeling},
 url = {https://proceedings.neurips.cc/paper_files/paper/2022/file/882d801fb1017f955547d5a816ade0fc-Paper-Conference.pdf},
 volume = {35},
 year = {2022}
}

@inproceedings{DBLP:conf/iclr/JongZFSC22,
  author       = {Michiel de Jong and
                  Yury Zemlyanskiy and
                  Nicholas FitzGerald and
                  Fei Sha and
                  William W. Cohen},
  title        = {Mention Memory: incorporating textual knowledge into Transformers
                  through entity mention attention},
  booktitle    = {The Tenth International Conference on Learning Representations, {ICLR}
                  2022, Virtual Event, April 25-29, 2022},
  publisher    = {OpenReview.net},
  year         = {2022},
  url          = {https://openreview.net/forum?id=OY1A8ejQgEX},
  bibsource    = {dblp computer science bibliography, https://dblp.org}
}

@inproceedings{DBLP:conf/emnlp/XiaoLBNSM22,
  author       = {Yuxin Xiao and
                  Paul Pu Liang and
                  Umang Bhatt and
                  Willie Neiswanger and
                  Ruslan Salakhutdinov and
                  Louis{-}Philippe Morency},
  editor       = {Yoav Goldberg and
                  Zornitsa Kozareva and
                  Yue Zhang},
  title        = {Uncertainty Quantification with Pre-trained Language Models: {A} Large-Scale
                  Empirical Analysis},
  booktitle    = {Findings of the Association for Computational Linguistics: {EMNLP}
                  2022, Abu Dhabi, United Arab Emirates, December 7-11, 2022},
  pages        = {7273--7284},
  publisher    = {Association for Computational Linguistics},
  year         = {2022},
  url          = {https://doi.org/10.18653/v1/2022.findings-emnlp.538},
  doi          = {10.18653/V1/2022.FINDINGS-EMNLP.538},
  bibsource    = {dblp computer science bibliography, https://dblp.org}
}

@inproceedings{lakshminarayanan2017simple,
 author = {Lakshminarayanan, Balaji and Pritzel, Alexander and Blundell, Charles},
 booktitle = {Advances in Neural Information Processing Systems},
 editor = {I. Guyon and U. Von Luxburg and S. Bengio and H. Wallach and R. Fergus and S. Vishwanathan and R. Garnett},
 pages = {},
 publisher = {Curran Associates, Inc.},
 title = {Simple and Scalable Predictive Uncertainty Estimation using Deep Ensembles},
 url = {https://proceedings.neurips.cc/paper_files/paper/2017/file/9ef2ed4b7fd2c810847ffa5fa85bce38-Paper.pdf},
 volume = {30},
 year = {2017}
}

@inproceedings{
malinin2020uncertainty,
title={Uncertainty Estimation in Autoregressive Structured Prediction},
author={Andrey Malinin and Mark Gales},
booktitle={International Conference on Learning Representations},
year={2021},
url={https://openreview.net/forum?id=jN5y-zb5Q7m}
}

@inproceedings{
DBLP:conf/iclr/KuhnGF23,
title={Semantic Uncertainty: Linguistic Invariances for Uncertainty Estimation in Natural Language Generation},
author={Lorenz Kuhn and Yarin Gal and Sebastian Farquhar},
booktitle={The Eleventh International Conference on Learning Representations },
year={2023},
url={https://openreview.net/forum?id=VD-AYtP0dve}
}

@article{DBLP:journals/tacl/FomichevaSYBGFA20,
    title = "Unsupervised Quality Estimation for Neural Machine Translation",
    author = "Fomicheva, Marina  and
      Sun, Shuo  and
      Yankovskaya, Lisa  and
      Blain, Fr{\'e}d{\'e}ric  and
      Guzm{\'a}n, Francisco  and
      Fishel, Mark  and
      Aletras, Nikolaos  and
      Chaudhary, Vishrav  and
      Specia, Lucia",
    editor = "Johnson, Mark  and
      Roark, Brian  and
      Nenkova, Ani",
    journal = "Transactions of the Association for Computational Linguistics",
    volume = "8",
    year = "2020",
    address = "Cambridge, MA",
    publisher = "MIT Press",
    url = "https://aclanthology.org/2020.tacl-1.35/",
    doi = "10.1162/tacl_a_00330",
    pages = "539--555"
}

@inproceedings{lin-etal-2022-towards,
    title = "Towards Collaborative Neural-Symbolic Graph Semantic Parsing via Uncertainty",
    author = "Lin, Zi  and
      Liu, Jeremiah Zhe  and
      Shang, Jingbo",
    editor = "Muresan, Smaranda  and
      Nakov, Preslav  and
      Villavicencio, Aline",
    booktitle = "Findings of the Association for Computational Linguistics: ACL 2022",
    month = may,
    year = "2022",
    address = "Dublin, Ireland",
    publisher = "Association for Computational Linguistics",
    url = "https://aclanthology.org/2022.findings-acl.328/",
    doi = "10.18653/v1/2022.findings-acl.328",
    pages = "4160--4173"
}

@inproceedings{DBLP:conf/acl/Chen024,
  author       = {Jiuhai Chen and
                  Jonas Mueller},
  editor       = {Lun{-}Wei Ku and
                  Andre Martins and
                  Vivek Srikumar},
  title        = {Quantifying Uncertainty in Answers from any Language Model and Enhancing
                  their Trustworthiness},
  booktitle    = {Proceedings of the 62nd Annual Meeting of the Association for Computational
                  Linguistics (Volume 1: Long Papers), {ACL} 2024, Bangkok, Thailand,
                  August 11-16, 2024},
  pages        = {5186--5200},
  publisher    = {Association for Computational Linguistics},
  year         = {2024},
  url          = {https://doi.org/10.18653/v1/2024.acl-long.283},
  doi          = {10.18653/V1/2024.ACL-LONG.283},
  bibsource    = {dblp computer science bibliography, https://dblp.org}
}

@article{DBLP:journals/tmlr/LinHE22,
title={Teaching Models to Express Their Uncertainty in Words},
author={Stephanie Lin and Jacob Hilton and Owain Evans},
journal={Transactions on Machine Learning Research},
issn={2835-8856},
year={2022},
url={https://openreview.net/forum?id=8s8K2UZGTZ},
note={}
}

@inproceedings{DBLP:conf/emnlp/BerantCFL13,    title = "Semantic Parsing on {F}reebase from Question-Answer Pairs",
    author = "Berant, Jonathan  and
      Chou, Andrew  and
      Frostig, Roy  and
      Liang, Percy",
    editor = "Yarowsky, David  and
      Baldwin, Timothy  and
      Korhonen, Anna  and
      Livescu, Karen  and
      Bethard, Steven",
    booktitle = "Proceedings of the 2013 Conference on Empirical Methods in Natural Language Processing",
    month = oct,
    year = "2013",
    address = "Seattle, Washington, USA",
    publisher = "Association for Computational Linguistics",
    url = "https://aclanthology.org/D13-1160/",
    pages = "1533--1544"
}

@inproceedings{DBLP:conf/nips/LinGOXLY024,
  author       = {Sheng{-}Chieh Lin and
                  Luyu Gao and
                  Barlas Oguz and
                  Wenhan Xiong and
                  Jimmy Lin and
                  Scott Yih and
                  Xilun Chen},
  editor       = {Amir Globersons and
                  Lester Mackey and
                  Danielle Belgrave and
                  Angela Fan and
                  Ulrich Paquet and
                  Jakub M. Tomczak and
                  Cheng Zhang},
  title        = {{FLAME} : Factuality-Aware Alignment for Large Language Models},
  booktitle    = {Advances in Neural Information Processing Systems 38: Annual Conference
                  on Neural Information Processing Systems 2024, NeurIPS 2024, Vancouver,
                  BC, Canada, December 10 - 15, 2024},
  year         = {2024},
  url          = {http://papers.nips.cc/paper\_files/paper/2024/hash/d16152d53088ad779ffa634e7bf66166-Abstract-Conference.html},
  bibsource    = {dblp computer science bibliography, https://dblp.org}
}

@inproceedings{sun-etal-2025-divide,
    title = "Divide-Then-Align: Honest Alignment based on the Knowledge Boundary of {RAG}",
    author = "Sun, Xin  and
      Xie, Jianan  and
      Chen, Zhongqi  and
      Liu, Qiang  and
      Wu, Shu  and
      Chen, Yuehe  and
      Song, Bowen  and
      Wang, Zilei  and
      Wang, Weiqiang  and
      Wang, Liang",
    editor = "Che, Wanxiang  and
      Nabende, Joyce  and
      Shutova, Ekaterina  and
      Pilehvar, Mohammad Taher",
    booktitle = "Proceedings of the 63rd Annual Meeting of the Association for Computational Linguistics (Volume 1: Long Papers)",
    month = jul,
    year = "2025",
    address = "Vienna, Austria",
    publisher = "Association for Computational Linguistics",
    url = "https://aclanthology.org/2025.acl-long.561/",
    doi = "10.18653/v1/2025.acl-long.561",
    pages = "11461--11480",
    ISBN = "979-8-89176-251-0"
}

@inproceedings{bi-etal-2025-context,
    title = "Context-{DPO}: Aligning Language Models for Context-Faithfulness",
    author = "Bi, Baolong  and
      Huang, Shaohan  and
      Wang, Yiwei  and
      Yang, Tianchi  and
      Zhang, Zihan  and
      Huang, Haizhen  and
      Mei, Lingrui  and
      Fang, Junfeng  and
      Li, Zehao  and
      Wei, Furu  and
      Deng, Weiwei  and
      Sun, Feng  and
      Zhang, Qi  and
      Liu, Shenghua",
    editor = "Che, Wanxiang  and
      Nabende, Joyce  and
      Shutova, Ekaterina  and
      Pilehvar, Mohammad Taher",
    booktitle = "Findings of the Association for Computational Linguistics: ACL 2025",
    month = jul,
    year = "2025",
    address = "Vienna, Austria",
    publisher = "Association for Computational Linguistics",
    url = "https://aclanthology.org/2025.findings-acl.536/",
    doi = "10.18653/v1/2025.findings-acl.536",
    pages = "10280--10300",
    ISBN = "979-8-89176-256-5"
}

@inproceedings{cohen-etal-2025-infact,
    title = "{I}n{F}act: Informativeness Alignment for Improved {LLM} Factuality",
    author = "Cohen, Roi  and
      Biswas, Russa  and
      de Melo, Gerard",
    editor = "Christodoulopoulos, Christos  and
      Chakraborty, Tanmoy  and
      Rose, Carolyn  and
      Peng, Violet",
    booktitle = "Findings of the Association for Computational Linguistics: EMNLP 2025",
    month = nov,
    year = "2025",
    address = "Suzhou, China",
    publisher = "Association for Computational Linguistics",
    url = "https://aclanthology.org/2025.findings-emnlp.971/",
    doi = "10.18653/v1/2025.findings-emnlp.971",
    pages = "17876--17888",
    ISBN = "979-8-89176-335-7"
}


\appendix


\section{Derivation of Rules for Mapping Uncertainty Values to Knowledge States}
\label{cognitive_state_mapping}

The proposed rule-based mapping from uncertainty values ($Consistency$ and $SE$) to natural-language knowledge states is as follow:

Let $Y_i = \{y_i^k\}_{k=1}^K$ be K sampled responses from model and $\hat{y}$ be the reference answer. 
\textit{Knowledge possession} is captured by $Consistency$, where:
\begin{itemize}
    \item $\mathrm{Consistency}>0$ indicates there exists at least one response $y_i^k$ matches the ground-truth answer $\hat{y}$, meaning the model possesses the required knowledge with a very high probability.
    \item $\mathrm{Consistency}=0$ represents that the model fails to answer the question correctly with $K$ times, indicating the model does not possess relevant knowledge to the question. 
    For a given question, we assume that if the LLM possesses the relevant knowledge, the probability of answering it correctly is $p_\theta = 0.5$. We set the confidence level to $\alpha=0.05$. If none of the $K$ sampled responses are correct, then with confidence $1-\alpha$ we can reject the hypothesis that the model’s probability of producing a correct answer satisfies $p_\theta \ge 0.5$. Under the assumption $p_\theta = 0.5$, sampling $K=6$ responses is sufficient to show that if none of them are correct, the model is unlikely to possess the relevant knowledge.
\end{itemize}

\textit{Answer Behavior} is measured by \textit{semantic entropy}:
\begin{itemize}
    \item The magnitude of semantic entropy reflects the model’s uncertainty at the semantic level: a higher value indicates diverse or conflicting semantic outputs (greater ambiguity), while a lower value suggests more consistent and deterministic semantic interpretations.
    \item Semantic entropy equals zero when all generated outputs are semantically equivalent, i.e., they fall into the same semantic cluster with no competing interpretations. From the semantic perspective, the model is completely certain, exhibiting neither ambiguity nor polysemy.
\end{itemize}

Given these interpretations, the mapping rules follow a logically consistent decision path:
\begin{enumerate}
    \item If $\mathrm{Consistency} > 0$ and $\mathrm{SE} = 0$, the model is judged to possess the relevant knowledge of a question and honestly provides consistent correct responses, corresponding to the knowledge state \textbf{KH}.
    \item If $\mathrm{Consistency} > 0$ and $\mathrm{SE} \neq 0$, the model produces a mix of correct and incorrect answers, indicating insufficient mastery of the knowledge to express it accurately. 
    The reason for this gap could be decoding strategy, hallucination snowballing, misalignment issues~\citep{liangLearningTrustYour2024}. 
    This corresponds to the knowledge state \textbf{K$\lnot$H}.
    \item If $\mathrm{Consistency} = 0$ and $\mathrm{SE} = 0$, the model lacks correct knowledge but converges on a single interpretation, corresponding to the knowledge state \textbf{$\lnot$KH}.
    \item In all other cases, the knowledge state is classified as \textbf{$\lnot$K$\lnot$H}.
\end{enumerate}

Overall, the mapping is determined by two factors:
\[
\underbrace{\text{Knowledge possession (Consistency)}}_{\text{know}}
\quad
\]
\text{and}
\[
\quad
\underbrace{\text{Answer honesty (Semantic Entropy)}}_{\text{tell}},
\]
which together define a quadrant of four cognitive states, ensuring both interpretability and completeness.

\section{Prompt Templates}
\label{prompt-template}
We illustrate the prompt templates used in this work in Figure~\ref{prompt}, detailing the input structure, incorporated knowledge states, and output format. 
The templates explicitly define how a question is combined with its corresponding knowledge state, optionally with retrieved external context, and then formatted to elicit model responses. By making this structure explicit, the figure clarifies how prompts guide the model during both training and inference, ensuring consistency across stages. Moreover, the design rationale highlights how natural-language descriptions of knowledge states are integrated into the prompt, which is essential for conveying uncertainty information in a semantically interpretable way.

\begin{figure}[htbp] 
\centering
    \begin{tcolorbox}[colframe=gray!50!black, colback=gray!5!white, title=Prompt template for sampling responses]
    \scriptsize
    
    You are an excellent Question-Answering assistant. Please answer the following question based on your knowledge.
    
    \vspace{3pt}
    \#\#\# Question \#\#\#: \{demo\_question\_1\}
    
    \#\#\# Answer \#\#\#: \{demo\_answer\_1\}

    \vspace{3pt}
    \#\#\# Question \#\#\#: \{input\_question\}
    
    \#\#\# Answer \#\#\#: 
    \end{tcolorbox}

    \begin{tcolorbox}[colframe=gray!50!black, colback=gray!5!white, title=Prompt template for supervised fine-tuning of the reference model]
    \scriptsize
    
    You are an excellent Question-Answering assistant. Please answer the following question based on your knowledge.
    
    \vspace{3pt}
    \#\#\# Question \#\#\#: \{THE QUESTION FROM DATASET\}
    
    \vspace{3pt}
    \#\#\# Self-Eval \#\#\#: \{THE KNOWLEDGE STATE FROM DATASET\}

    \vspace{3pt}
    \#\#\# Output \#\#\#: \{GOLDEN ANSWER\}
    \end{tcolorbox}

    \begin{tcolorbox}[colframe=gray!50!black, colback=gray!5!white, title=Prompt template for policy model optimization]
    \scriptsize
    
    You are an excellent Question-Answering assistant. Please answer the following question based on your knowledge.
    
    \vspace{3pt}
    \#\#\# Question \#\#\#: \{THE QUESTION FROM DATASET\}
    
    \vspace{3pt}
    \#\#\# Self-Eval \#\#\#: \{THE KNOWLEDGE STATE FROM DATASET\}

    \vspace{3pt}
    \#\#\# Answer \#\#\#: \{GOLDEN ANSWER\}

    \end{tcolorbox}

    \begin{tcolorbox}[colframe=gray!50!black, colback=gray!5!white, title=Prompt template for RAFT a RAG model]
    \scriptsize
    
    You are an excellent Question-Answering assistant. Please answer the following question based on your knowledge.
    
    \vspace{3pt}
    \#\#\# Question \#\#\#: \{THE QUESTION FROM DATASET\}
    
    \vspace{3pt}
    \#\#\# Self-Eval \#\#\#: \{THE KNOWLEDGE STATE FROM DATASET\}

    \vspace{3pt}
    \#\#\# Prior Judgment \#\#\#: \{RANDOMLY SELECTED RESPONSE FROM $Y_i$\}

    \vspace{3pt}
    \#\#\# Retrieve Documents \#\#\#: related passages: \#\#\#passage 1\#\#\#;\#\#\#passage 2\#\#\#;\#\#\#passage 3\#\#\#

    \vspace{3pt}
    \#\#\# Posterior Answer \#\#\#: \{GOLDEN ANSWER\}
    \end{tcolorbox}

    \begin{tcolorbox}[colframe=gray!50!black, colback=gray!5!white, title=Prompt template for RAG model in inference]
    \scriptsize
    
    You are an excellent Question-Answering assistant. Please answer the following question based on your knowledge.
    
    \vspace{3pt}
    \#\#\# Question \#\#\#: \{THE QUESTION FROM DATASET\}
    
    \vspace{3pt}
    \#\#\# Self-Eval \#\#\#: \{THE KNOWLEDGE STATE FROM DATASET\}

    \vspace{3pt}
    \#\#\# Prior Judgment \#\#\#: \{POLICY MODEL'S OUTPUT\}

    \vspace{3pt}
    \#\#\# Retrieve Documents \#\#\#: related passages: \#\#\#passage 1\#\#\#;\#\#\#passage 2\#\#\#;\#\#\#passage 3\#\#\#

    \vspace{3pt}
    \#\#\# Posterior Answer \#\#\#: \{GOLDEN ANSWER\}
    \end{tcolorbox}

    \begin{tcolorbox}[colframe=gray!50!black, colback=gray!5!white, title=Prompt template for supervised fine-tuning of the estimator model.]
    \scriptsize
    
    You are an excellent Question-Answering assistant. Please answer the following question based on your knowledge.
    
    \vspace{3pt}
    \#\#\# Question \#\#\#: \{THE QUESTION FROM DATASET\}
    
    \vspace{3pt}
    \#\#\# Self-Eval \#\#\#: \{THE KNOWLEDGE STATE FROM DATASET\}

    \end{tcolorbox}
    
\caption{All the prompt templates employed in FAITH (our framework).}
\label{prompt}
\end{figure}

\section{Details of Training}
\label{training_details}
All experiments are conducted on a cluster equipped with 4 × NVIDIA A40 and/or 4 × NVIDIA 4090D GPUs. 

For supervised fine-tuning (SFT) of both the reference model and estimator in FAITH, we train for 3 epochs. We adopt the Adam optimizer with an initial learning rate of 2e-4. We apply LoRA with a rank of 32, alpha of 16, and a dropout rate of 0.05, targeting all layers. The batch size per device is set to 8, with gradient accumulation steps of 8, leading to a total batch size of 256. The learning rate scheduler follows a cosine decay with a warmup ratio of 0.0. 

For policy optimization, both the reward model (RM) and the PPO stages are trained for 2 epochs. We adopt the Adam optimizer with an initial learning rate of 1e-5. LoRA is applied with a rank of 8, alpha of 16, and a dropout rate of 0.05, again targeting all layers. The per-device batch size is set to 4, with gradient accumulation steps of 8, leading to a total batch size of 128. The learning rate scheduler is cosine decay, and the warmup ratio is 0.0, consistent with the SFT stage.

\section{Details of Dataset}
\label{dataset detail}
\paragraph{SciQ:}The SciQ dataset~\citep{DBLP:conf/aclnut/WelblLG17} contains 13,679 crowdsourced science examination questions covering subjects such as physics, chemistry, and biology. Although originally released in multiple-choice format, in our setting all answer options are removed, and each question is reformulated as an open-ended query requiring a direct answer. For most questions, an accompanying paragraph with supporting evidence is provided, offering factual context that can be utilized to guide answer generation and factual alignment. In our experiments, 11,679 samples are used for training and 1,000 samples are reserved for validation, with the remaining questions serving as an in-domain test set.
\paragraph{TriviaQA:} TriviaQA~\citep{joshi-etal-2017-triviaqa} is a large-scale reading comprehension dataset containing over 650K question-answer-evidence triples, with questions authored by trivia enthusiasts and evidence documents collected from Wikipedia and the web. In our work, for constructing the augmented dataset, we pre-process and sample half of the original training set. 

\paragraph{NQ-Open:} NQ-Open~\citep{kwiatkowski-etal-2019-natural} is an open-domain QA benchmark derived from the Natural Questions dataset, where real user queries are paired with English Wikipedia passages as the knowledge source. In our work, we employ NQ-Open for augmented dataset construction. Similarly, to ensure fair comparison and reduce computational cost, we sample half of the original training data.

\paragraph{Web-Questions:} The WebQuestions dataset~\citep{DBLP:conf/emnlp/BerantCFL13} comprises 6,642 question–answer pairs, where each question can be answered using Freebase, a large-scale knowledge graph. The majority of questions are centered around a single named entity and reflect typical queries collected from the web around 2013. In our experiments, we only employ its test set (1348 item) for the evaluation under the out-of-domain evaluation setting.

\section{Details of Evaluation Metrics}
\label{evaluation metric}
Truthfulness quantifies the proportion of correct responses among all provided answers, reflecting the LLM’s overall reliability in expressing knowledge. The formula for Truthfulness is given as follows:
\begin{equation}
\text{Truthfulness} = \frac{\text{UR} + \text{KC}}{\text{KC} + \text{KI} + \text{KR} + \text{UC} + \text{UI} + \text{UR}}
\label{eq:truthfulness}
\end{equation}

Precision measures the proportion of correctly answered questions among those for which the model possesses the relevant knowledge, reflecting the LLM’s ability to accurately convey known facts. The formula for Precision is given as follows:
\begin{equation}
\text{Precision} = \frac{\text{KC}}{\text{KI} + \text{KC} + \text{KR}}
\label{eq:precision}
\end{equation}

\section{Numerical Results of Training-time Scaling}
\label{app:numerical results}
Table~\ref{tab:self-eval-K-results} reports the detailed numerical results corresponding to the training-time scaling analysis. The table compares precision and truthfulness across different values of \(K\) (\(6, 8, 10, 12\)). Consistent with Figure~\ref{fig:training_scaling}, the results show that increasing \(K\) beyond 6 does not yield noticeable gains, confirming that \(K=6\) is sufficient to capture the model’s knowledge state distribution during data augmentation.

\begin{table*}[htbp]
  \centering
  \resizebox{\linewidth}{!}{
    \begin{tabular}{ccccccccccc}
    \toprule
    \multicolumn{1}{c}{\multirow{2}[4]{*}{\textbf{\# Responses}}} & \multicolumn{2}{c}{\textbf{TVQA (ID)}} & \multicolumn{2}{c}{\textbf{SciQ (ID)}} & \multicolumn{2}{c}{\textbf{NQ-Open (ID)}} & \multicolumn{2}{c}{\textbf{Average (ID)}}& \multicolumn{2}{c}{\textbf{WebQ-QA (OOD)}} \\
\cmidrule{2-11}          & \multicolumn{1}{c}{\textit{Prec.} $\uparrow$} & \multicolumn{1}{c}{\textit{Truth.} $\uparrow$} & \multicolumn{1}{c}{\textit{Prec.} $\uparrow$} & \multicolumn{1}{c}{\textit{Truth.} $\uparrow$} & \multicolumn{1}{c}{\textit{Prec.} $\uparrow$} & \multicolumn{1}{c}{\textit{Truth.} $\uparrow$} & \multicolumn{1}{c}{\textit{Prec.} $\uparrow$} & \multicolumn{1}{c}{\textit{Truth.} $\uparrow$}& \multicolumn{1}{c}{\textit{Prec.} $\uparrow$} & \multicolumn{1}{c}{\textit{Truth.} $\uparrow$} \\ 
\specialrule{0.1em}{0pt}{0pt} 

        \rowcolor[rgb]{0.91, 0.91, 0.91} \multicolumn{11}{c}{\rowstrut Llama-3-8B} \\

\specialrule{0pt}{0ex}{.05ex}  
    \multicolumn{1}{c}{\textbf{K=6}} 
    &82.95 &59.80 &80.29 &49.70 &57.99 &26.52 &73.79 &45.69 &67.31 &33.75
    \\
    \multicolumn{1}{c}{\textbf{K=8}}  & 84.36 & 63.55 & 81.52 & 54.70 & 60.95 & 25.51 & 75.61 & 47.92 &  67.42 & 37.76 \\
    \multicolumn{1}{c}{\textbf{K=10}} & 84.01  & 63.79   & 80.94  & 55.20 & 60.44  &  25.98 & 75.13 & 48.32  &  67.10  & 38.43 \\
    \multicolumn{1}{c}{\textbf{K=12}} & 84.36  &  64.65  & 80.76 & 55.40 & 60.35  &  26.40 & 75.16 & 48.82  & 66.15 & 38.28 \\
    \bottomrule
    \end{tabular}%
    }
\caption{Training-time scaling with different numbers of sampled responses ($K$) on Llama-3-8B.}
  \label{tab:self-eval-K-results}
\end{table*}%

\begin{table*}[htbp]
  \centering
  \small
  \resizebox{\linewidth}{!}{
    \begin{tabular}{ccccccccccc}
    \toprule
    \multicolumn{1}{c}{\multirow{2}[4]{*}{\textbf{Method}}} & \multicolumn{2}{c}{\textbf{TVQA (ID)}} & \multicolumn{2}{c}{\textbf{SciQ (ID)}} & \multicolumn{2}{c}{\textbf{NQ-Open (ID)}} & \multicolumn{2}{c}{\textbf{Average (ID)}}& \multicolumn{2}{c}{\textbf{WebQ-QA (OOD)}} \\
\cmidrule{2-11}          & \multicolumn{1}{c}{\textit{Prec.} $\uparrow$} & \multicolumn{1}{c}{\textit{Truth.} $\uparrow$} & \multicolumn{1}{c}{\textit{Prec.} $\uparrow$} & \multicolumn{1}{c}{\textit{Truth.} $\uparrow$} & \multicolumn{1}{c}{\textit{Prec.} $\uparrow$} & \multicolumn{1}{c}{\textit{Truth.} $\uparrow$} & \multicolumn{1}{c}{\textit{Prec.} $\uparrow$} & \multicolumn{1}{c}{\textit{Truth.} $\uparrow$}& \multicolumn{1}{c}{\textit{Prec.} $\uparrow$} & \multicolumn{1}{c}{\textit{Truth.} $\uparrow$} \\ 
\specialrule{0.1em}{0pt}{0pt} 
    \rowcolor[rgb]{0.91, 0.91, 0.91} \multicolumn{11}{c}{\rowstrut Llama3-8B} \\

\specialrule{0pt}{0ex}{.05ex}  
    \multicolumn{1}{l}{\textbf{FAITH$_{sample-base}$}} &  \textbf{85.02} &\textbf{60.98} &\textbf{81.43} &\textbf{50.33} &\textbf{59.25} &\textbf{28.02} &\textbf{75.23} &\textbf{46.44} &\textbf{68.67} &\textbf{34.89}                  \\
    \multicolumn{1}{l}{\textbf{FAITH$_{Estimator}$}} &  84.19  &  60.69  &  80.61  &  49.99    & 58.13 & 27.58 & 74.26  &  45.73    & 67.99     &    34.03            \\
        \rowcolor[rgb]{0.91, 0.91, 0.91} \multicolumn{11}{c}{\rowstrut Mistral-7B-v0.1} \\

\specialrule{0pt}{0ex}{.05ex}  
    \multicolumn{1}{l}{\textbf{FAITH$_{sample-base}$}} &   
    \textbf{88.12} &\textbf{61.43} &\textbf{82.53} &\textbf{52.23} &\textbf{48.97} &\textbf{24.59} &\textbf{73.21} &\textbf{46.08} &\textbf{58.87} &\textbf{41.23}          \\
    \multicolumn{1}{l}{\textbf{FAITH$_{Estimator}$}} & 87.20 & 60.72 & 81.42 & 51.40 & 48.05 & 23.91 & 72.22  & 45.34   &  58.04  &  40.43   \\
    \bottomrule
    \end{tabular}%
    }
\vspace{-5pt}
\caption{Performance Comparison of FAITH\_sample-base and FAITH\_Estimator.}

  \label{tab:estimater-eval-results}
\end{table*}%

\section{Contribution of Estimator}
We have added new results to Table \ref{tab:estimater-eval-results} (shown below) comparing FAITH$_{sample-base}$ (using $K=6$ responses) and FAITH$_{Estimator}$. The results show that while the sampling-based method provides a slightly higher upper bound, our Estimator achieves comparable performance (within ~1\% difference) while significantly reducing inference latency by avoiding multiple sampling rounds.

\section{Case Study}
\label{app:case_study}
In our method FAITH, one focus is to train RAG model to align the policy model's output with external knowledge. The RAG model is provided with retrieved passages as context, allowing it to rectify or retain the policy model’s responses. In this section, we analyze three types of corrections, with representative cases shown in Table~\ref{case study1}, \ref{case study2}, and~\ref{case study3}, as case studies. Specifically, the three correction types are summarized as follows:
\begin{enumerate}
    \item \textbf{Implicitly Supported Correction}:
    The initial answer from the policy model was incorrect, but after applying our trained RAG model, the final answer was corrected. Notably, the retrieved passages did not verbatim reproduce the correct answer, but contained key information or semantic cues related to the correct answer. Details can be found in Table \ref{case study1}.
    \item \textbf{Explicitly Supported Correction}:
    The policy model initially produced an incorrect output, but after applying our trained RAG model, the final output was corrected. In this process, the retrieved content from RAG not only directly reproduced the correct answer but also provided additional information related to it, thereby supporting the model’s correction. Details can be found in Table \ref{case study2}.
    \item \textbf{Misleading Override}:
    The policy model initially produced the correct answer. However, after applying our trained RAG model, the output was incorrectly altered. This occurred because the retrieved content contained misleading information that contradicted the correct answer, ultimately leading to an erroneous output.  Details can be found in Table \ref{case study3}.
\end{enumerate}

\section{Reproducibility Statement}
We have made several efforts to ensure the reproducibility of our work. We provided detailed descriptions of the datasets used in our experiments, all of which are publicly available. Our method is thoroughly explained in dedicated sections~\S~\ref{methods}, and we also provide detailed training parameters~\S~\ref{subsec:exp_setup}. Finally, we have publicized the code at \url{https://github.com/xndong/FAITH}.

We hope that these measures will facilitate the replication of our work by other researchers and further advance the field.

\section{The Use of Large Language Model}
The authors acknowledge the use of OpenAI ChatGPT solely for enhancing the coherence of the final manuscript, and providing assistance with coding for data processing.

\begin{table*}[tp]
\small
\centering
\setlength{\tabcolsep}{6pt} 
\renewcommand{\arraystretch}{1.3} 
\begin{tabular}{m{1cm}|p{3cm}| p{5cm}|p{4cm}}
\toprule
\textbf{Type} & \multicolumn{1}{c|}{\textbf{Question \& Answer}} & \multicolumn{1}{c|}{\textbf{Retrieved Passages}} & \multicolumn{1}{c}{\textbf{Our Analysis}} \\
\midrule
\multirow{34}{*}{\centering\rotatebox{90}{\textbf{Implicitly Supported Correction}}}&\textbf{Q:} Protists play critically important ecological roles as producers and, on the other end of food webs, as what? \newline\ \textbf{A:} {decomposers}\newline \textbf{A1:} consumers \newline \textbf{A2:} decomposers
    & \textbf{I.} In real world ecosystems, there is more than one food chain for most organisms, since most organisms eat more than one kind of food or are eaten by more than one type of predator. A diagram that sets out the intricate network of intersecting and overlapping food chains for an ecosystem is called its food web. Decomposers are often left off food webs, but if included, they mark the end of a food chain. \textbf{Thus, food chains start with primary producers and end with decay and decomposers.} \newline \textbf{II.} Food webs have trophic levels and positions. Basal species, such as plants, form the first level and are the resource-limited species that feed on no other living creature in the web. Basal species can be autotrophs or detritivores, including decomposing organic material and its associated microorganisms, which we defined as detritus, micro-inorganic material, and associated microorganisms (MIP), and vascular plant material. \newline \textbf{III.} The microbial food web refers to the combined trophic interactions among microbes in aquatic environments. These microbes include viruses, bacteria, algae, and heterotrophic protists (such as ciliates and flagellates). In aquatic environments, microbes constitute the base of the food web. Single-celled photosynthetic organisms such as diatoms and cyanobacteria are generally the most important primary producers in the open ocean. Many of these cells, especially cyanobacteria, are too small to be captured and consumed by small crustaceans and planktonic larvae. Instead, these cells are consumed by phagotrophic protists, which are readily consumed by larger organisms.
&  Some protists do function as A1 (consumers), but the model fails to accurately address the specific context of “the other end of food webs” posed in the question. This suggests the model's insufficiency in effectively utilizing its internal knowledge to answer the question, or a failure to correctly understand the question, particularly its key constraints. After introducing the retrieved information to our trained RAG model, it successfully corrected the answer to A2 (decomposers). The retrieved information, in paragraph I, provides the crucial background knowledge: “Thus food chains start with primary producers and end with decay and decomposers.” This information does not explicitly state that “protists are decomposers.” Instead, it requires the model to synthesize this information with “protists” and “the other end of food webs” to deduce the correct answer. We denote this process as \textbf{Implicitly Supported Correction}.
\\
\bottomrule
\end{tabular}
\caption{Case Study Analysis 1 of Model Responses. Specifically, \textbf{A} denotes Ground Truth, \textbf{A1} denotes the policy model's output, and  \textbf{A2} denotes the final output.}
\label{case study1}
\end{table*}

\begin{table*}[ht]
\small
\centering
\setlength{\tabcolsep}{6pt} 
\renewcommand{\arraystretch}{1.3} 
\begin{tabular}{m{1cm}|p{3cm}| p{5cm} |p{4cm}}
\toprule
\textbf{Type} & \multicolumn{1}{c|}{\textbf{Question \& Answer}} & \multicolumn{1}{c|}{\textbf{Retrieved Passages}} & \multicolumn{1}{c}{\textbf{Our Analysis}} \\
\midrule
\multirow{34}{*}{\centering\rotatebox{90}{\textbf{Explicitly Supported Correction}}}&\textbf{Q:} Rita Coolidge sang the title song for which Bond film? \newline\ \textbf{A:} {Octopussy} \newline \textbf{A1:} North by Northwest \newline \textbf{A2:} Octopussy
& \textbf{I.} Octopussy is the soundtrack for the eponymous thirteenth James Bond film. The score was composed by John Barry, the lyrics by Tim Rice. \textbf{The opening theme, All Time High is sung by Rita Coolidge and is one of six Bond film title songs or songs that are not named after the film's title.} The original compact disc released in 1985 by A\&M Records, was recalled because of a printing error and became a rarity.  \newline \textbf{II.} Another Way to Die is a song by American musicians Jack White and Alicia Keys. Written and produced by White as the theme song to the 2008 James Bond film Quantum of Solace, it was released as a single in the United States on September 30, 2008, and in Europe on October 20, 2008. \newline \textbf{III.} Tomorrow Never Dies is the song, performed by Sheryl Crow, which served as the theme song to the James Bond film of the same name. The song was co-written by Crow and the song's producer Mitchell Froom, and became her fifth UK Top 20 hit, peaking at No. 12 in 1997. Another song, Tomorrow Never Dies, written by the movie's composer David Arnold and performed by k.d. lang, was originally produced as the official theme tune. When Crow's song became the official theme, the k.d. lang song was relegated to the end credits, and renamed Surrender.
&  The model's initial response, A1 (North by Northwest), is a significant factual error, as this film is not even part of the James Bond series. This indicates a substantial knowledge gap or a “hallucination” in the model's internal knowledge base. After the RAG intervention, the model successfully corrected the answer to A2 (Octopussy). The retrieved information in paragraph I contains all the key details required to rectify the error. The passage explicitly states, “The opening theme, All Time High is sung by Rita Coolidge” and that the soundtrack was for the film Octopussy. The model simply needed to match the key entity from the question, “Rita Coolidge”, with the retrieved text to directly find the name of the film for which she sang the theme. The entire process involves direct information extraction and localization, requiring almost no complex reasoning. We denote this process as \textbf{Explicitly Supported Correction}.
\\
\bottomrule
\end{tabular}
\caption{Case Study Analysis 2 of Model Responses. Specifically, \textbf{A} denotes Ground Truth, \textbf{A1} denotes the policy model's output, and  \textbf{A2} denotes the final output.}
\label{case study2}
\end{table*}

\begin{table*}[ht]
\small
\centering
\setlength{\tabcolsep}{6pt} %
\renewcommand{\arraystretch}{1.3} %
\begin{tabular}{m{1cm}|p{3cm}| p{5cm} |p{4cm}}
\toprule
\textbf{Type} & \multicolumn{1}{c|}{\textbf{Question \& Answer}} & \multicolumn{1}{c|}{\textbf{Retrieved Passages}} & \multicolumn{1}{c}{\textbf{Our Analysis}} \\
\midrule
\multirow{34}{*}{\centering\rotatebox{90}{\textbf{Misleading Override}}}&\textbf{Q:} Which grand slam did Pete Sampras not win in the 20th century? \newline\ \textbf{A:} {French} \newline \textbf{A1:} French Open \newline \textbf{A2:} Wimbledon
& \textbf{I.} As the Swiss national anthem played Federer was overcome with emotion after finally capturing the elusive title at Roland Garros. This match was momentous in the history of tennis. After missing the chance to equal Pete Sampras' then-record of fourteen Grand Slam championships of all time when he lost to Rafael Nadal in the final of the Australian Open earlier in the year, Federer finally did so by winning the French Open for the first time. Sampras himself commented on Federer following the victory saying, Regardless he [Federer] goes down as the greatest ever.\newline \textbf{II.} In the eight Wimbledons inclusive between 1993 and 2000, 1996 was the only year that Sampras would fail to win the championship at Wimbledon. Sampras lost in the quarterfinals of Wimbledon to the eventual winner, Richard Krajicek, the tournament's 17th-seed. The match lasted three long sets, with Krajicek winning 7-5, 7-6, 6-4. In the quarterfinals of the US Open, Sampras vomited on the court at 1–1 in the final set tiebreak (due to dehydration) while facing Àlex Corretja; nonetheless, Sampras would win that match. \newline \textbf{III.} He beat former champion Michael Stich in the fourth round and met Sampras in the quarterfinals. By that time, he had managed to turn his notably weak slice backhand into an aggressive top-spin shot. Krajicek shocked the tennis world by defeating Sampras in straight sets, becoming the only player to beat Sampras in a Wimbledon singles match in the eight-year period from 1993 until Sampras' fourth-round loss to Roger Federer in the 2001 tournament.
&  This is a failure case of a 'correct-to-incorrect' reversal caused by the RAG. The model's initial judgment, A1 (French Open), was correct, indicating that its internal knowledge base already contained the key fact about Sampras's career. However, the intervention of RAG instead led to a degradation in performance. The core of the failure lies in the Retrieval stage. The retrieved information, though related to the key entities “Pete Sampras” and “Grand Slam“, did not align with the question's specific requirement (“did not win” in the 20th century). The retrieved content, particularly in paragraphs II and III, repeatedly and in detail described a specific loss Sampras had at Wimbledon (in 1996 to Krajicek). Phrases like “fail to win the championship at Wimbledon” became a strong and irrelevant distracting signal. When generating the final answer, the model over-relied on this incorrectly retrieved and distracting content, thereby ignoring its own correct prior knowledge. It was misled into outputting the incorrect answer A2 (Wimbledon). We denite this process as \textbf{Misleading Override}.
\\
\bottomrule
\end{tabular}
\caption{Case Study Analysis 3 of Model Responses. Specifically, \textbf{A} denotes Ground Truth, \textbf{A1} denotes the policy model's output, and \textbf{A2} denotes the final output.}
\label{case study3}
\end{table*}

\end{document}